\documentclass[letterpaper, 10 pt, conference]{ieeeconf}  

\IEEEoverridecommandlockouts                              

\usepackage{times}

\usepackage[table, svgnames]{xcolor}
\usepackage[pagebackref=true,breaklinks=true,letterpaper=true,colorlinks,bookmarks=false]{hyperref}
\usepackage{url}
\usepackage{booktabs}       
\usepackage{amsfonts}       
\usepackage{nicefrac}       
\usepackage{microtype}      
\usepackage{color}
\usepackage{xspace}
\usepackage{graphicx}
\usepackage{caption}
\usepackage{subcaption}
\usepackage{array}
\usepackage{diagbox}
\usepackage{wrapfig}
\usepackage{anyfontsize}
\newcommand{\PreserveBackslash}[1]{\let\temp=\\#1\let\\=\temp}
\newcolumntype{C}[1]{>{\PreserveBackslash\centering}p{#1}}
\newcolumntype{R}[1]{>{\PreserveBackslash\raggedleft}p{#1}}
\newcolumntype{L}[1]{>{\PreserveBackslash\raggedright}p{#1}}

\makeatletter
\newcommand*{\@rowstyle}{}

\newcommand*{\rowstyle}[1]{
  \gdef\@rowstyle{\leavevmode#1}%
  \@rowstyle\ignorespaces%
}

\newcolumntype{=}{>{\gdef\@rowstyle{}}}
\newcolumntype{+}{>{\@rowstyle}}

\makeatother
\usepackage{amssymb}
\usepackage{amsmath,amsfonts,amssymb}
\usepackage{amsopn}
\usepackage{bm} 
\usepackage{multirow}

\newcommand{\ProbOpr}[1]{\mathbb{#1}}

\newcommand{\expect}[2]{%
\ifthenelse{\equal{#2}{}}{\ProbOpr{E}_{#1}}
{\ifthenelse{\equal{#1}{}}{\ProbOpr{E}\left[#2\right]}{\ProbOpr{E}_{#1}\left[#2\right]}}} 
\newcommand{\var}[2]{%
\ifthenelse{\equal{#2}{}}{\ProbOpr{VAR}_{#1}}
{\ifthenelse{\equal{#1}{}}{\ProbOpr{VAR}\left[#2\right]}{\ProbOpr{VAR}_{#1}\left[#2\right]}}} 

\newcommand{\eat}[1]{}
\newcommand{\method}[1]{\textsc{#1}}

\newcommand{\PIXOR}{\method{PIXOR}\xspace}

\newcommand{\APBEV}{AP$_\text{BEV}$\xspace}
\newcommand{\AP}{AP$_\text{3D}$\xspace}

\DeclareRobustCommand{\eg}{e.g.\@\xspace}
\DeclareRobustCommand{\ie}{i.e.\@\xspace}

\newcommand{\dreaming}{\method{Dreaming}\xspace}

\title{\LARGE \bf
Exploiting Playbacks in Unsupervised Domain Adaptation \\for 3D Object Detection in Self-Driving Cars
}

\author{Yurong You$^{*\dag}$, Carlos Andres Diaz-Ruiz$^{*\ddag}$, Yan Wang$^\dag$, Wei-Lun Chao$^\S$, \\
Bharath Hariharan$^\dag$, Mark Campbell$^\ddag$, Kilian Q Weinberger$^\dag$%
\thanks{$*$ Equal contributions}%
\thanks{$\dag$ Computer Science Department, Cornell University 
        {\tt \{yy785, yw763, bh497, kqw4\}@cornell.edu}}%
\thanks{$\ddag$ Mechanical and Aerospace Engineering Department, Cornell University 
        {\tt \{cad297, mc288\}@cornell.edu}}%
\thanks{$\S$ Department of Computer Science and Engineering, the Ohio State University 
        {\tt chao.209@osu.edu}}
}

\newcommand{\lyft}{Lyft\xspace}
\newcommand{\argo}{Argoverse\xspace}
\newcommand{\nusc}{nuScenes\xspace}
\newcommand{\waymo}{Waymo\xspace}
\newcommand{\kitti}{KITTI\xspace}

\begin{document}

\maketitle
\thispagestyle{empty}
\pagestyle{empty}


\begin{abstract}
Self-driving cars must detect other traffic participants like vehicles and pedestrians in 3D in order to plan safe routes and avoid collisions. 
State-of-the-art 3D object detectors, based on deep learning, have shown promising accuracy but are prone to over-fit domain idiosyncrasies, making them fail in new environments---a serious problem for the robustness of self-driving cars. 
In this paper, we propose a novel learning approach that reduces this gap by fine-tuning the detector on high-quality pseudo-labels in the target domain -- pseudo-labels that are automatically generated {\textit{after driving}} based on replays of previously recorded driving sequences. In these replays, object tracks are smoothed forward and backward in time, and detections are interpolated and extrapolated---crucially, leveraging future information to catch hard cases such as missed detections due to occlusions or far ranges. We show, across five autonomous driving datasets, that fine-tuning the object detector on these pseudo-labels substantially reduces the domain gap to new driving environments, yielding strong improvements detection reliability and accuracy.
\end{abstract}


\section{Introduction}

Detecting traffic participants such as cars, cyclists, and pedestrians in 3D is a fundamental learning problem for self-driving cars.
Typically, inputs consists of LiDAR point clouds and/or images; outputs are sets of tight 3D bounding boxes that envelop  detected objects. The problem is challenging, because the detection must be highly accurate and reliable.
The current state of the art in 3D object detection is based on deep learning approaches \cite{qi2018frustum,shi2019pointrcnn,yang2018pixor,shi2020pv}, trained on short driving segments with labeled bounding boxes \cite{geiger2012we,geiger2013vision}, which yield up to $80\%$ average precision on held-out segments \cite{shi2020pv}.

However, as with all machine learning approaches, these techniques succeed when the training and test data distributions match.
One way to ensure this is to constrain self-driving cars to a small geo-fenced area, such as with a fleet of similar self-driving taxis collecting and sharing training data about the same area. 
This approach, however, is not generalizable to consumer self-driving cars which 
should be able to drive freely anywhere, similar to a human-driven car. 
This unconstrained scenario introduces an inherent adaptation problem: the car producer cannot foresee where the owner will ultimately operate the car. For example, the perception system might be trained on urban roads in Germany \cite{geiger2012we,geiger2013vision}, but the car may be driven in the mountain roads in the USA, where cars are larger and fewer, roads may be snowier, and the environment (trees, roads, etc.) may look different.
Past work has shown that such differences can cause  $>\!35\%$ drop in detection accuracy \cite{yan2020domain}.
Closing this adaptation gap is one of the biggest challenges for consumer self-driving vehicles. 

Formally, this challenge is a problem in \emph{unsupervised domain adaptation (UDA)}~\cite{gong2012geodesic}: 
the detector, having been previously trained on labeled data from a \emph{source} domain, must now adapt to a \emph{target} domain where only unlabeled data are available. While the UDA problem is easily cast with training and testing datasets in different locations (Germany vs USA, urban vs rural, etc.), similar problems exist in many self-driving car scenarios. For example, consider the common case of car owners who drive many of the same routes (\eg, commuting), and leave their cars parked (\eg, at night) for extended periods of time. This raises an intriguing possibility: the car can collect sensor data on these trips, and then retrain itself autonomously \emph{offline} to adapt to this new environment, to improve subsequent \emph{online} driving.

In this paper, we present a novel approach for UDA of 3D detectors for self-driving cars. Our approach uniquely uses two key insights. First, data collected  over time via a \emph{video} is not simply a bag of independent frames. Second, the \emph{dynamics} of our objects of interest (\ie, cars) can be modeled effectively. We propose to use time correlations between frames and object physics to enable more accurate and efficient solutions to the UDA problem {\em offline}. More specifically, 
our approach takes the `confident' detections of nearby objects, estimates their states (\eg, locations, sizes and speeds), and then \emph{extrapolates} the tracks forward and backward in time, discovering challenging cases when the detector made mistakes (\eg missed detections due to occlusions, detections at far ranges, etc.) 
The \emph{playback} of the data allows us to go back in time and annotate labels in frames which were previously missed. 
Although this process cannot be performed in real time (since it uses future information), we can use it \emph{offline} to generate a new training set with pseudo-labels for the target environment. We can then adapt the detector to the target domain using this newly created dataset, thus allowing the detector to generalize to more scenarios.
We call our approach \emph{dreaming}, as the car learns by replaying past driving sequences backwards and forwards, potentially while it is parked overnight. 

We evaluate our approach on the most challenging of UDA problems, that of 3D object detection across multiple autonomous driving datasets, including \kitti \cite{geiger2012we,geiger2013vision}, \argo \cite{argoverse}, \lyft \cite{lyft2019}, \waymo \cite{waymo_open_dataset}, and \nusc \cite{nuscenes2019}.
We show \emph{across all dataset combinations} that discovering detector mistakes and retraining the detector with our dreaming car procedure strongly reduces the source/target domain gap with high consistency. 
In fact, the resulting detector after ``dreaming'' \emph{substantially exceeds}  the accuracy of the offline system used to generate the pseudo-labels, which---although able to look into the future---is limited to the extrapolation of confident detections before adaption. Our dreaming procedure can easily be implemented on-device and we believe that it constitutes a significant step towards safely operating autonomous vehicles without geo-restrictions. 

\section{Related Work}
\label{sec:related}

\textbf{3D object detection} can be categorized based on the input: using 3D sensors 
like LiDAR or 2D images from cameras \cite{pseudoLiDAR,you2019pseudo,E2EPL,li2019stereo}. We focus on the former due to its higher accuracy.
Examples of LiDAR-based detectors include F-PointNet \cite{qi2018frustum}, PointRCNN \cite{shi2019pointrcnn}, PIXOR \cite{yang2018pixor} and VoxelNet \cite{zhou2018voxelnet}.
While these methods have consistently improved the detection accuracy, it has been recently revealed in \cite{yan2020domain} that they cannot generalize well when trained and tested on different datasets, especially on distant objects with sparse LiDAR points. 

\textbf{Unsupervised domain adaptation (UDA)}
has been widely studied in machine learning and computer vision, especially on image classification~\cite{gong2012geodesic,ganin2016domain,saito2018maximum}.  The common setup is to adapt a model trained from one labeled source domain (\eg, synthetic images) to another unlabeled target domain (\eg, real images).
Recent work has extended UDA to driving scenes, 
but mainly for 2D object detection \cite{chen2018domain,he2019multi,kim2019diversify,saito2019strong,zhu2019adapting} and semantic segmentation \cite{hoffman2017cycada,saleh2018effective,tsai2018learning,zhang2019curriculum,zou2018unsupervised}.
The mainstream approach is to match the feature distributions or image appearances between domains, e.g., via adversarial learning \cite{ganin2016domain, hoffman2017cycada} or image translation \cite{zhu2017unpaired}. 
The approaches more similar to ours are \cite{roychowdhury2019automatic,tao2018zero,liang2019distant,zhang2018collaborative,zou2018unsupervised,kim2019self,french2018self,inoue2018cross,khodabandeh2019robust}, which iteratively assign pseudo-labels to (some of) the unlabeled data in the target domain and retrain the models. 
This procedure, usually named self-training, is proven to be effective in learning with unlabeled data, such as semi-supervised and weakly-supervised learning~\cite{mcclosky2006effective,kumar2020understanding,lee2013pseudo,cinbis2016weakly}. For UDA, self-training enables the model to adapt its features to the target domain in a supervised fashion.

\textbf{UDA in 3D} tackles the domain discrepancy in point clouds.
Qin et al. \cite{qin2019pointdan} were the first to match point cloud distributions between domains, via adversarial learning. 
However, they considered point clouds of isolated objects, which are very different from the ones captured in driving scenes.
Other approaches project 3D points to the frontal or bird's-eye view and apply UDA methods in the resulting 2D images
\cite{saleh2019domain,wang2019range,wu2019squeezesegv2}, which can be sub-optimal in models' accuracy.

Instead, \emph{we follow the self-training paradigm and show that UDA in 3D object detection can be drastically improved by our high-quality pseudo-labels.} Concurrent work ST3D~\cite{yang2021st3d} also tackles UDA in 3D via self-training, but it focuses on addressing the object size discrepancy across domains~\cite{yan2020domain}. It pre-trains a 3D detector with random object scaling and uses pseudo-labels to train student detectors in a curriculum-learning way. In contrast, our work makes use of consecutive LiDAR scans in the target domain to improve pseudo-label quality, especially for faraway or previously missed objects. Our work is thus orthogonal and complementary to theirs.

\textbf{Leveraging videos for object detection} has been explored in \cite{liang2015towards,ovsep2019large,misra2015watch,kumar2016track} to ease the labeling efforts by mining extra 2D bounding boxes from videos.
The main idea is to leverage the temporal information 
to extend weakly-labeled instances or potential object proposals across frames, which are then used as pseudo-labels to retrain the detectors. In the context of UDA, \cite{tang2012shifting,roychowdhury2019automatic,ovsep2017large} also incorporate object tracks to discover high quality 2D pseudo-labels for self-training.

Our approach is different in two aspects. First, we not only interpolate but also \emph{extrapolate} tracks to infer object locations when they are too far away to be detected accurately. 
Second, we operate in 3D. 
We apply a physical-based motion model and exploit the fact that objects in 3D are scale-invariant to correct the detection along tracks.
In contrast, the methods above operate in 2D and may disregard faraway objects (that appear too small in images) due to unreliable 2D tracks \cite{ovsep2017large}.

\textbf{Auto-labeling}. Our work is also related to concurrent work in 3D auto-labeling~\cite{qi2021offboard,yang2021auto4d}, which
improves the initial detections of a 3D object detector in an offline manner by aggregating them across the whole sequence with a standard tracker~\cite{weng20203d}. Our work is different from theirs in two aspects. First, our approach not only aggregates detections in a track, but \emph{extrapolates} them with a motion model, which is crucial to recover the missing detections faraway (\autoref{tbl:k2a_ablation}). Second, our end-goal is to improve the 3D object detector by fine-tuning it with the improved pseudo-labels, while auto-labeling solely focuses on improving pseudo-labels.

\section{Exploiting Playbacks for UDA}

Similar to most published work on 3D object detection for autonomous driving \cite{qi2018frustum,shi2019pointrcnn,yang2018pixor,shi2020pv,lang2019pointpillars}, we focus on \emph{frame-wise} 3D detectors.   
A detector is first trained on a source domain and then applied to a target domain (\eg, a new city).
\cite{yan2020domain} revealed a drastic accuracy drop in such a scenario: many of the target objects are either misdetected or mislocalized, especially if they are far away. 
To aid adaptation, we assume access to an unlabeled dataset of \emph{video sequences} in the target domain, which could simply be recordings of the vehicle's sensors while it was in operation. 
Our approach is to generate pseudo-labels for these recordings that can be used to adapt the detector to the new environment during periods when the car is not in use. 
We assume no access to the source data when performing adaptation---it is unlikely that car producers share data with the customers after the detector is deployed.

\subsection{Tracking for improved detection}
\label{sub-sec:tracking}
One way to improve the test accuracy based on the frame-wise detection outputs is \emph{online} tracking by detection~\cite{breitenstein2010online,breitenstein2009robust,hua2015online}.  Here,  detected objects are associated across current and \emph{past} frames to derive  trajectories, which are used to filter out false positives, correct false negatives, and adjust the initial detection boxes in the current frame.

\textbf{Online 3D object tracking.}  We investigate this idea with a Kalman filter based tracker~\cite{8843260,chiu2020probabilistic,weng20203d}, which has shown promising results in benchmark tracking leader boards~\cite{nuscenes2019}. \emph{We opt to not use a learning-based tracker~\cite{yin2020center} since it would also require adaptation before it can be applied in the target domain.} 
Specifically, we apply the tracker in \cite{8843260}. The algorithm estimates the joint probability $p(\mathbf{a}_k,\mathbf{x}_k|\mathbf{z}_k)$ at time $k$, where $\mathbf{x}_k$ is the set of tracked object states (\eg, cars speeds and locations), $\mathbf{z}_k$ is the set of observed sensor measurements (here each measurement is a frame-wise detection), and $\mathbf{a}_k$ is the assignment of measurements to tracks. The joint probability can be decomposed into continuous estimation $p(\mathbf{x}_k|\mathbf{a}_k,\mathbf{z}_k)$ and discrete data assignment $p(\mathbf{a}_k|\mathbf{z}_k)$.
The former is solved recursively via an Extended Kalman Filter (EKF); the latter is solved via Global Nearest-Neighbor (GNN), where the cost to be minimized is the negative BEV IoU between the predicted box and the measurement. The EKF parameterizes the state $\mathbf{x}$ of a single ($i$th) object as a vehicle (position, velocity, shape) \textit{relative} to the ego-vehicle
\begin{equation}
\mathbf{x}^{i}_{k}=
    \begin{bmatrix}
    x & y & \theta & s & l & w
    \end{bmatrix}^T,
\end{equation}
where $x$, $y$ are the location of the tracked vehicle's back axle relative to a fixed point on the ego-vehicle, $\theta$ is the vehicle orientation relative to the ego-vehicle, $s$ is the absolute ground speed, and $l, w$ are the length and width. 
The EKF uses a dynamics model of the evolution of the state over time. Here we assume that the tracked vehicle is moving at a \emph{constant} speed and heading in the global coordinate frame, with added noise to represent the uncertainty associated with vehicle maneuvers.
This tracker has been shown to work well on tracking moving objects from a self-driving car~\cite{8843260, 5643163}.
We initialize a new track when $c_{\mathrm{min-hits}}$ measurements of the same tracked object are realized.
We end a track when it does not obtain any measurement updates over $c_{\mathrm{max-age}}$ frames, or the tracked object exits the field of view (FOV).
As will be seen in~\autoref{ssec:results}, applying this tracker can indeed improve the detection accuracy online, via inputting missed detections, correcting mislocalized detections, and rejecting wrong detections in $\mathbf{z}_k$ at current time $k$ by $\mathbf{x}_k$.

\textbf{Offline 3D object tracking.} Online trackers only use past information to improve current detections. By relaxing this for offline tracking (\eg, to be able to look into the future and come back to the current time), we can obtain more accurate estimates of vehicle states. 
While such a relaxation is not applicable during test time, higher accuracy tracking on unlabeled driving sequences will be very valuable for adapting the source detector to the target domain in a self-supervised fashion, as we will explain in the following section.

\subsection{Self-training for UDA}
Self-training is a simple yet fairly effective way to improve a model with unlabeled data~\cite{mcclosky2006effective,kumar2020understanding,lee2013pseudo,chen2011co}. 
The basic idea is to apply an existing model to an unlabeled dataset  
and use the high confidence predictions (here detections), which are likely to be correct, as ``pseudo-labels'' for fine-tuning.
One key to success for self-training is the quality of the pseudo-labels.
In particular, we desire two qualities out of the detections we use as pseudo-labels.
First, they should be \emph{correct}, i.e., they should not include false positives.
Second, they should have \emph{high coverage}, i.e., they should cover all cases of objects. 
Choosing high confidence detections as pseudo-labels satisfies the first criterion but not the second. For 3D object detection, we find that most of the high confidence examples are easy cases: unoccluded objects near the self-driving car. This is where offline tracking becomes a crucial component to include the more challenging cases (far away, or partially occluded objects) in the pseudo-label pool. 

\subsection {High quality pseudo-labels via 3D offline tracking}
\label{ssec:PL_T}

\begin{figure*}[t]
    \vspace{5pt} 
    \centering
    \includegraphics[width=.85\textwidth]{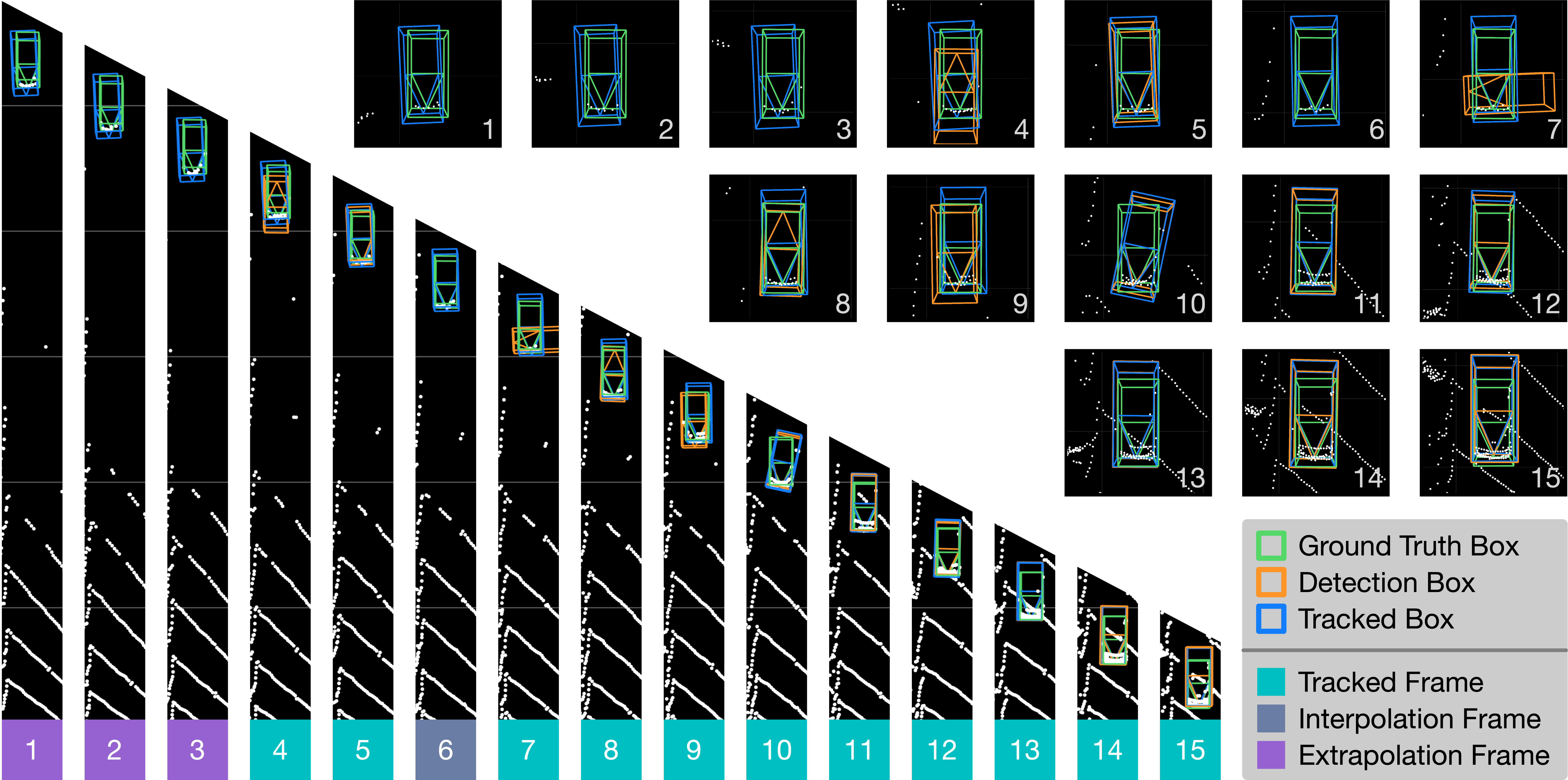}
    \caption{An example tracked vehicle moves towards ego-vehicle (at the bottom), where pseudo-labels are recovered through extrapolation, interpolation, and smoothing. The improvements of pseudo-labels via offline tracking are instances where estimated bounding boxes (blue) are observed, while the frame-wise detections (orange) are missing or poorly aligned with ground truths (green). Better viewed in color.
}
    \label{fig:overview}
    \vspace{-15pt}
\end{figure*}
How do we obtain pseudo-labels for far-away, hard-to-detect objects that the detector cannot reliably detect?
We propose to exploit tracking by leveraging two facts in the autonomous driving scenario.
First, the available unlabeled data is in the form of \emph{sequences} (akin to videos) of point clouds over time.
Second, the objects of interest and the self-driving car move in fairly constrained ways.
We will run the object detector on \emph{logged} data, so that we can easily analyze both forwards and backwards in time.
The object detector will detect objects accurately only when they are close to the self-driving car.
Once detected over a few frames, we can estimate the object's motion either towards the self-driving car or away from it, and then both \emph{interpolate} the object's positions in frames where it was missed, or \emph{extrapolate} the object into frames where it is too far away for accurate detection.
We show an example of this procedure in
 \autoref{fig:overview}. Through dynamic modeling, tracking,  and smoothing over time we can
correct noisy detections. Further with extrapolation and interpolation, we can recover far away and missed detections.

Concretely, we proposed to augment the following functionalities that utilize the future information into the online tracker introduced in~\autoref{sub-sec:tracking}, turning it into an offline tracker specifically designed to improve detection, \ie, generating higher quality pseudo-labels for self-training. 

\textbf{State smoothing.} 
Frame-wise 3D object detectors can generate inconsistent, noisy detection across time (e.g., frame 4, 7 and 9 in \autoref{fig:overview}).
The model-based tracking approach in~\autoref{sub-sec:tracking} reduces this noise, but we can go further by smoothing tracks back and forth over time, since our data is offline.
In this work, we use a fixed-point Rauch-Tung-Striebel (RTS) smoother \cite{bar2001estimation} to smooth the tracked state estimates. 
Smoothing requires a backward iteration ($k=N,N-1,...1)$ that is performed after
the forward filtering, where the a-posteriori state and state error covariance
estimates ($\bar{\mathbf{x}}_{k|k}$ and $ P_{k|k}$) and the a-priori state and state error covariance
estimates ($\bar{\mathbf{x}}_{k+1|k}$ and $P_{k+1|k}$) have been calculated. The smoothed gain, $C_{k}$, is obtained from
\begin{equation}
 C_{k} = P_{k|k}F_{k}^{T}P_{k+1|k}^{-1},
\end{equation}
where $F_{k}$ is the Jacobian of the dynamics model evaluated at $\bar{\mathbf{x}}_{k|k}$.
The smoothed state is then evaluated as
\begin{equation}
    \bar{\mathbf{x}}_{k|N} = \bar{\mathbf{x}}_{k|k} +C_{k}[\bar{\mathbf{x}}_{k+1|N}-\bar{\mathbf{x}}_{k|k}]
\end{equation}
while the covariance of the smoothed state is evaluated as
\begin{equation}
    P_{k|N} = P_{k|k} + C_{k}[P_{k+1|N}-P_{k+1|k}]C_{k}^{T}.
\end{equation}

\textbf{Adjusting object sizes.} 
As shown in \cite{yan2020domain}, the distribution of car sizes in different domains (e.g. different cities) can be different.
As such, when tested on a novel domain, detectors often predict incorrect object sizes.
This is especially true when the LiDAR signal is too sparse for correct size information.
We can also use our tracking to correct such systematic error.
Assuming that the most confident detections are more likely to be accurate, we estimate the size of the object by averaging the size of the three highest confidence detections.
We use this size for all objects in this track.

\textbf{Interpolation and extrapolation.} 
We use estimation (forward in time) and smoothing (backward in time) to recover missed detections, and in turn, to increase the
recall rate of pseudo-labels (e.g., frame 1--3 and 6 in \autoref{fig:overview}). If a detection is missed in the middle of a track,
we restore it by taking the estimated state from  smoothing. We also extrapolate
the tracks both backward and forward in time, so that tracks that were
prematurely terminated due to missing detections, can be recovered. 
More concretely,
we
are able to recover detections of vehicles that were lost as they moved away from the ego-vehicle because the sensor signals became sparser (or in turn, the vehicles started far away and were then only detected when they got close enough). Extrapolations are performed by first using dynamics model predictions of the EKF to predict potential bounding boxes;  measurements are obtained by performing a search and detection in the vicinity of the prediction. We apply
the detector in a 3 m$^2$ area around the extrapolated prediction, yielding several 3D bounding box candidates. After filtering out candidates with confidences lower than some threshold, we select the candidate with the highest BEV IoU with the prediction as the measurement. If a track loses such a measurement for three consecutive frames, extrapolations are stopped. With this targeted search, we are able to recover objects that were missed due to low confidence. After extrapolating and interpolating detections for all tracks, we perform Non Maximum Suppression (NMS) over bounding boxes in BEV, where more recent extrapolations/interpolations are prioritized.

\textbf{Discussion.} The tracker we apply is standard and simple. We opt it to show the power of our dreaming approach for UDA---exploiting offline, forward and backward information to derive high-quality pseudo-labels for adapting detectors. More
sophisticated trackers will likely improve the results further.
While we focus on frame-wise 3D detectors, our algorithm can be applied to adapt video-based 3D object detectors \cite{yin2020lidar} as well. One particular advantage of fine-tuning on the pseudo-labeled target data is to allow the detector adapting not only its predictions (\eg, the box regression) but also its features (\eg, early layers in the neural networks) to the target domain. The resulting detector thus can usually lead to more accurate detections than the pseudo-labels it has been trained on.

\section{Experiments}
\label{sec:exp}
\textbf{Datasets.} We experiment with five autonomous driving data sets: \kitti \cite{geiger2012we,geiger2013vision}, \argo \cite{argoverse}, \nusc \cite{nuscenes2019}, \lyft \cite{lyft2019} and \waymo \cite{waymo_open_dataset}. All datasets provide LiDAR data in sequences and ground-truth bounding box labels for either all or part of the data. We briefly summarize these five datasets in the supplementary. We follow a setup similar to \cite{yan2020domain}, but with different splits on \lyft, \nusc, and \waymo in order to keep the training and test sets non-overlapping with sequences. For \nusc and \waymo, we only use data from a single location (Boston for \nusc and San Francisco for \waymo).

\textbf{UDA settings.} We train models in the source domain using labeled frames. We split each dataset into two non-overlapping parts, a training set and a test set. \emph{We use the \textbf{train} set (and its pseudo-labels) of the target domain to adapt the source detector, and evaluate the adapted detector on the \textbf{test} set.}

\textbf{Metric.}
We follow \kitti to evaluate object detection in 3D and BEV metrics. We focus on the \emph{Car} category as it is the main focus of existing work. We report average precision (AP) with the intersection over union (IoU) thresholds at 0.5 or 0.7, \ie, a car is correctly detected if the IoU between it and the detected box is larger than 0.5 or 0.7. We denote AP in 3D and BEV by \AP and \APBEV, respectively. Because on the other datasets there is no official separation on the difficulty levels like in \kitti, we split AP by depth ranges.

\textbf{3D object detection models.} We use two LiDAR-based models \method{PointRCNN}~\cite{shi2019pointrcnn} and \PIXOR~\cite{yang2018pixor} to detect objects in 3D. They represent two different but popular ways of processing point cloud data. 
\method{PointRCNN} uses PointNet++~\cite{qi2017pointnet++} to extract point-wise features, while \PIXOR applies 2D convolutions in BEV of voxelized point clouds. Neither relies on images. We mainly report and discuss results of \method{PointRCNN} except for the last study in \autoref{ssec:results}.

\textbf{Hyper-parameters.} To train detectors on source domain, we use the hyper-parameters provided by \cite{shi2019pointrcnn} for \method{PointRCNN}. For \PIXOR, we follow \cite{yan2020domain} to train it using RMSProp with momentum $0.9$ and learning rate $5\times 10^{-5}$ (decreased by a factor of 10 after 50 and 80 epochs) for 90 epochs. 
For self-training on the target domain, we initialize from the pre-trained model on the source domain. For \method{PointRCNN}, we fine-tune it with learning rate $2\times 10^{-4}$ and 40 epochs in RPN and 10 epochs in RCNN. For \PIXOR, we use RMSProp with momentum $0.9$ and learning rate $5 \times 10^6$ (decreased by a factor of 10 after 10 and 20 epochs) for 30 epochs.

We developed and tuned our dreaming method with \argo as the source domain and \kitti as the target domain (in the target domain, we only use the training set).
We then fixed all hyper-parameters for all subsequent experiments. 

\subsection{Baselines}
\label{ssec:baselines}
We compare against two baselines under the UDA setting.

\textbf{Self-Training (ST).} We apply a self-training scheme similar to that typically used in the 2D problems~\cite{kumar2020understanding}. When adapting the model from the source to the target, we apply the source model to the target training set. We then keep the detected cars of confidence scores $>0.8$ (label-sharpening) and use them as pseudo-labels to fine-tune the model. We select the threshold following our hyper-parameter selection procedure and apply it to all the experiments.

\textbf{Statistical Normalization (SN).} \cite{yan2020domain} showed that car sizes vary between domains: popular cars at different areas can be different.
When the mean bounding box size in the target domain is accessible, either from limited amount of labeled data or statistical data, we can apply \emph{statistical normalization (SN)}~\cite{yan2020domain} to mitigate such a systematic difference in car sizes. SN adjusts the bounding box sizes and corresponding point clouds in the source domain to match those in the target domain, and fine-tunes the model on such ``normalized'' source data, with no need to access target sensor data. We follow the exact setting in \cite{yan2020domain} to apply SN.
\subsection{Empirical Results}
\label{ssec:results}

\begin{table}[!bt]
    \vspace{5pt} 
	\tabcolsep 1.2pt
	\fontsize{7}{8}\selectfont
	\renewcommand{\arraystretch}{1.05}
	\centering
	\caption{\textbf{Pseudo-label quality.} We compare the quality of the pseudo-labels on \kitti \textbf{train} set generated by a detector trained on \argo (\textsc{PL}) and those after smoothing, resizing, interpolation, and extrapolation (\textsc{PL (after)}). We show the \APBEV{}/ \AP of the \textbf{car} category at IoU $= 0.7$ and $0.5$ across different depth range. \label{tbl:pseudo_label_quality}}
	\vspace{-.5em}
    \begin{tabular}{=l|+c|+c|+c|+c|+c|+c}
    \multicolumn{1}{c|}{} & \multicolumn{3}{c|}{IoU 0.5} & \multicolumn{3}{c}{IoU 0.7} \\ \cline{2-7}
    \multicolumn{1}{c|}{\multirow{-2}{*}{Method}}            & \color{black}0-30    & \color{black}30-50    & \color{black}50-80   & \color{black}0-30    & \color{black}30-50    & \color{black}50-80   \\ \hline
    \textsc{PL} & \textbf{80.5} / 80.0 & 63.7 / 60.7 & 23.9 / 18.6 & 63.9 / 36.6 & 33.0 / 12.1 & 10.0 / \textbf{\phantom{0}0.9} \\ 
    \textsc{PL (after)} & \textbf{80.5 / 80.2} & \textbf{68.0 / 61.1} & \textbf{29.5 / 22.6} & \textbf{64.9 / 38.0} & \textbf{38.4 / 15.1} & \textbf{12.1 / \phantom{0}0.9}\\
    \hline
    \end{tabular}
\vspace{-20pt}
\end{table}

\textbf{Pseudo-label quality.} In \autoref{tbl:pseudo_label_quality} we evaluate the quality of pure pseudo-labels and pseudo-labels after smoothing, resizing, interpolation, and extrapolation under the \argo to \kitti setting. It can be seen that the dreaming process significantly improves the pseudo-label quality across all ranges, which leads to further improvement after self-training.


\begin{table*}[ht!]
\vspace{3pt} 
\tabcolsep 1.5pt
\fontsize{6.45}{7.5}\selectfont
\renewcommand{\arraystretch}{1.05}
\caption{\textbf{Dreaming results on UDA among five auto-driving datasets.} We report \APBEV and \AP of the Car category on far-away range (50-80m) and full range (0-80m) at IoU$=0.5$. On each entry (\emph{row}, \emph{column}), we report AP of UDA from \emph{row} to \emph{column} in the order of No Re-training / Self-Training / Dreaming. At the diagonal entries is the AP of in-domain model. Our method is marked in {\color{blue}blue}. \label{tbl:5x5}}
\vspace{-.5em}
\begin{subtable}{.49\textwidth}
\caption{50 -- 80 m}
\begin{tabular}{r|c|c|c|c|c}
    \APBEV & \multicolumn{1}{c|}{\kitti} & \multicolumn{1}{c|}{\argo} & \multicolumn{1}{c|}{\lyft} & \multicolumn{1}{c|}{\nusc} & \multicolumn{1}{c}{\waymo} \\ \hline
    \kitti & {\color{gray}40.4}&20.1\hspace{0.1em}/\hspace{0.1em}26.9\hspace{0.1em}/\hspace{0.1em}{\color{blue}\textbf{28.4}}&49.6\hspace{0.1em}/\hspace{0.1em}56.3\hspace{0.1em}/\hspace{0.1em}{\color{blue}\textbf{56.4}}&\phantom{0}1.4\hspace{0.1em}/\hspace{0.1em}\textbf{\phantom{0}9.1}\hspace{0.1em}/\hspace{0.1em}{\color{blue}\phantom{0}4.5}&42.8\hspace{0.1em}/\hspace{0.1em}48.5\hspace{0.1em}/\hspace{0.1em}{\color{blue}\textbf{50.2}}\\
    \argo  &18.0\hspace{0.1em}/\hspace{0.1em}28.2\hspace{0.1em}/\hspace{0.1em}{\color{blue}\textbf{30.1}}& {\color{gray}37.8}&46.3\hspace{0.1em}/\hspace{0.1em}48.8\hspace{0.1em}/\hspace{0.1em}{\color{blue}\textbf{54.5}}&\phantom{0}0.6\hspace{0.1em}/\hspace{0.1em}\textbf{\phantom{0}9.1}\hspace{0.1em}/\hspace{0.1em}{\color{blue}\phantom{0}3.0}&50.4\hspace{0.1em}/\hspace{0.1em}50.9\hspace{0.1em}/\hspace{0.1em}{\color{blue}\textbf{56.0}}\\
    \lyft  &26.1\hspace{0.1em}/\hspace{0.1em}30.8\hspace{0.1em}/\hspace{0.1em}{\color{blue}\textbf{33.9}}&29.3\hspace{0.1em}/\hspace{0.1em}30.7\hspace{0.1em}/\hspace{0.1em}{\color{blue}\textbf{35.2}}& {\color{gray}67.2}&\phantom{0}4.5\hspace{0.1em}/\hspace{0.1em}\textbf{\phantom{0}9.1}\hspace{0.1em}/\hspace{0.1em}{\color{blue}\phantom{0}6.0}&51.6\hspace{0.1em}/\hspace{0.1em}51.9\hspace{0.1em}/\hspace{0.1em}{\color{blue}\textbf{56.9}}\\
    \nusc  &\phantom{0}9.6\hspace{0.1em}/\hspace{0.1em}16.4\hspace{0.1em}/\hspace{0.1em}{\color{blue}\textbf{21.7}}&\phantom{0}3.0\hspace{0.1em}/\hspace{0.1em}12.4\hspace{0.1em}/\hspace{0.1em}{\color{blue}\textbf{17.7}}&22.9\hspace{0.1em}/\hspace{0.1em}39.1\hspace{0.1em}/\hspace{0.1em}{\color{blue}\textbf{46.4}}& {\color{gray}\phantom{0}3.5}&24.9\hspace{0.1em}/\hspace{0.1em}42.5\hspace{0.1em}/\hspace{0.1em}{\color{blue}\textbf{50.1}}\\
    \waymo &14.3\hspace{0.1em}/\hspace{0.1em}25.6\hspace{0.1em}/\hspace{0.1em}{\color{blue}\textbf{27.8}}&24.7\hspace{0.1em}/\hspace{0.1em}23.4\hspace{0.1em}/\hspace{0.1em}{\color{blue}\textbf{28.3}}&45.3\hspace{0.1em}/\hspace{0.1em}54.0\hspace{0.1em}/\hspace{0.1em}{\color{blue}\textbf{55.5}}&\phantom{0}0.2\hspace{0.1em}/\hspace{0.1em}\phantom{0}3.0\hspace{0.1em}/\hspace{0.1em}{\color{blue}\textbf{\phantom{0}9.1}}& {\color{gray}58.2}\\\hline
\end{tabular}\\
\begin{tabular}{r|c|c|c|c|c}
    \AP & \multicolumn{1}{c|}{\kitti} & \multicolumn{1}{c|}{\argo} & \multicolumn{1}{c|}{\lyft} & \multicolumn{1}{c|}{\nusc} & \multicolumn{1}{c}{\waymo} \\ \hline
    \kitti & {\color{gray}36.3} & 15.4\hspace{0.1em}/\hspace{0.1em}19.5\hspace{0.1em}/\hspace{0.1em}{\color{blue}\textbf{20.8}} & 40.0\hspace{0.1em}/\hspace{0.1em}46.3\hspace{0.1em}/\hspace{0.1em}{\color{blue}\textbf{47.7}} & \textbf{\phantom{0}0.9}\hspace{0.1em}/\hspace{0.1em}\phantom{0}0.4\hspace{0.1em}/\hspace{0.1em}{\color{blue}\phantom{0}0.4} & 33.4\hspace{0.1em}/\hspace{0.1em}39.3\hspace{0.1em}/\hspace{0.1em}{\color{blue}\textbf{41.0}} \\
    \argo   & 13.9\hspace{0.1em}/\hspace{0.1em}22.5\hspace{0.1em}/\hspace{0.1em}{\color{blue}\textbf{25.6}}& {\color{gray}30.0} & 42.9\hspace{0.1em}/\hspace{0.1em}46.0\hspace{0.1em}/\hspace{0.1em}{\color{blue}\textbf{47.6}} & \phantom{0}0.1\hspace{0.1em}/\hspace{0.1em}\phantom{0}0.2\hspace{0.1em}/\hspace{0.1em}{\color{blue}\textbf{\phantom{0}3.0}} & 47.8\hspace{0.1em}/\hspace{0.1em}48.2\hspace{0.1em}/\hspace{0.1em}{\color{blue}\textbf{49.1}} \\
    \lyft   & 20.4\hspace{0.1em}/\hspace{0.1em}24.0\hspace{0.1em}/\hspace{0.1em}{\color{blue}\textbf{26.4}} & 25.3\hspace{0.1em}/\hspace{0.1em}26.7\hspace{0.1em}/\hspace{0.1em}{\color{blue}\textbf{27.3}}& {\color{gray}65.5} & \phantom{0}0.4\hspace{0.1em}/\hspace{0.1em}\textbf{\phantom{0}9.1}\hspace{0.1em}/\hspace{0.1em}{\color{blue}\phantom{0}1.1} & 49.6\hspace{0.1em}/\hspace{0.1em}50.6\hspace{0.1em}/\hspace{0.1em}{\color{blue}\textbf{50.8}} \\
    \nusc   & \phantom{0}5.5\hspace{0.1em}/\hspace{0.1em}\phantom{0}8.7\hspace{0.1em}/\hspace{0.1em}{\color{blue}\textbf{13.4}} & \phantom{0}3.0\hspace{0.1em}/\hspace{0.1em}\phantom{0}9.1\hspace{0.1em}/\hspace{0.1em}{\color{blue}\textbf{13.0}} & 15.1\hspace{0.1em}/\hspace{0.1em}30.2\hspace{0.1em}/\hspace{0.1em}{\color{blue}\textbf{37.8}}& {\color{gray}\phantom{0}2.8} & 22.5\hspace{0.1em}/\hspace{0.1em}39.6\hspace{0.1em}/\hspace{0.1em}{\color{blue}\textbf{45.5}} \\
    \waymo  & \phantom{0}8.5\hspace{0.1em}/\hspace{0.1em}18.1\hspace{0.1em}/\hspace{0.1em}{\color{blue}\textbf{19.7}} & 19.6\hspace{0.1em}/\hspace{0.1em}21.3\hspace{0.1em}/\hspace{0.1em}{\color{blue}\textbf{22.3}} & 43.0\hspace{0.1em}/\hspace{0.1em}48.1\hspace{0.1em}/\hspace{0.1em}{\color{blue}\textbf{49.6}} & \phantom{0}0.0\hspace{0.1em}/\hspace{0.1em}\phantom{0}0.3\hspace{0.1em}/\hspace{0.1em}{\color{blue}\textbf{\phantom{0}9.1}}& {\color{gray}50.8} \\\hline
\end{tabular}
\end{subtable}
\hfill
\begin{subtable}{.49\linewidth}
\caption{0 -- 80 m}
\begin{tabular}{r|c|c|c|c|c}
    \APBEV & \multicolumn{1}{c|}{\kitti} & \multicolumn{1}{c|}{\argo} & \multicolumn{1}{c|}{\lyft} & \multicolumn{1}{c|}{\nusc} & \multicolumn{1}{c}{\waymo} \\ \hline
    \kitti & {\color{gray}87.1}&56.8\hspace{0.1em}/\hspace{0.1em}58.8\hspace{0.1em}/\hspace{0.1em}{\color{blue}\textbf{59.4}}&68.2\hspace{0.1em}/\hspace{0.1em}68.3\hspace{0.1em}/\hspace{0.1em}{\color{blue}\textbf{68.6}}&27.7\hspace{0.1em}/\hspace{0.1em}27.8\hspace{0.1em}/\hspace{0.1em}{\color{blue}\textbf{28.9}}&62.5\hspace{0.1em}/\hspace{0.1em}68.8\hspace{0.1em}/\hspace{0.1em}{\color{blue}\textbf{69.0}}\\
    \argo  &82.3\hspace{0.1em}/\hspace{0.1em}83.2\hspace{0.1em}/\hspace{0.1em}{\color{blue}\textbf{83.3}}& {\color{gray}68.5}&66.7\hspace{0.1em}/\hspace{0.1em}67.0\hspace{0.1em}/\hspace{0.1em}{\color{blue}\textbf{67.6}}&\textbf{26.9}\hspace{0.1em}/\hspace{0.1em}25.2\hspace{0.1em}/\hspace{0.1em}{\color{blue}25.4}&69.8\hspace{0.1em}/\hspace{0.1em}69.6\hspace{0.1em}/\hspace{0.1em}{\color{blue}\textbf{70.0}}\\
    \lyft  &82.6\hspace{0.1em}/\hspace{0.1em}84.9\hspace{0.1em}/\hspace{0.1em}{\color{blue}\textbf{85.3}}&60.2\hspace{0.1em}/\hspace{0.1em}65.7\hspace{0.1em}/\hspace{0.1em}{\color{blue}\textbf{66.0}}& {\color{gray}79.2}&28.8\hspace{0.1em}/\hspace{0.1em}28.2\hspace{0.1em}/\hspace{0.1em}{\color{blue}\textbf{29.1}}&70.6\hspace{0.1em}/\hspace{0.1em}70.9\hspace{0.1em}/\hspace{0.1em}{\color{blue}\textbf{71.1}}\\
    \nusc  &61.7\hspace{0.1em}/\hspace{0.1em}76.5\hspace{0.1em}/\hspace{0.1em}{\color{blue}\textbf{79.5}}&22.4\hspace{0.1em}/\hspace{0.1em}38.0\hspace{0.1em}/\hspace{0.1em}{\color{blue}\textbf{46.6}}&41.1\hspace{0.1em}/\hspace{0.1em}59.0\hspace{0.1em}/\hspace{0.1em}{\color{blue}\textbf{65.5}}& {\color{gray}37.7}&51.5\hspace{0.1em}/\hspace{0.1em}62.2\hspace{0.1em}/\hspace{0.1em}{\color{blue}\textbf{68.7}}\\
    \waymo &81.2\hspace{0.1em}/\hspace{0.1em}82.2\hspace{0.1em}/\hspace{0.1em}{\color{blue}\textbf{83.1}}&56.9\hspace{0.1em}/\hspace{0.1em}58.3\hspace{0.1em}/\hspace{0.1em}{\color{blue}\textbf{59.2}}&67.3\hspace{0.1em}/\hspace{0.1em}68.9\hspace{0.1em}/\hspace{0.1em}{\color{blue}\textbf{69.4}}&23.1\hspace{0.1em}/\hspace{0.1em}27.2\hspace{0.1em}/\hspace{0.1em}{\color{blue}\textbf{29.1}}& {\color{gray}71.9}\\\hline
\end{tabular}
\begin{tabular}{r|c|c|c|c|c}
    \AP & \multicolumn{1}{c|}{\kitti} & \multicolumn{1}{c|}{\argo} & \multicolumn{1}{c|}{\lyft} & \multicolumn{1}{c|}{\nusc} & \multicolumn{1}{c}{\waymo} \\ \hline
    \kitti & {\color{gray}86.7} & 49.0\hspace{0.1em}/\hspace{0.1em}55.0\hspace{0.1em}/\hspace{0.1em}{\color{blue}\textbf{55.9}} & 60.7\hspace{0.1em}/\hspace{0.1em}65.2\hspace{0.1em}/\hspace{0.1em}{\color{blue}\textbf{66.4}} & 20.9\hspace{0.1em}/\hspace{0.1em}22.3\hspace{0.1em}/\hspace{0.1em}{\color{blue}\textbf{23.1}} & 59.4\hspace{0.1em}/\hspace{0.1em}60.5\hspace{0.1em}/\hspace{0.1em}{\color{blue}\textbf{61.4}} \\
    \argo   & 77.4\hspace{0.1em}/\hspace{0.1em}\textbf{81.1}\hspace{0.1em}/\hspace{0.1em}{\color{blue}\textbf{81.1}}& {\color{gray}65.9} & 64.9\hspace{0.1em}/\hspace{0.1em}65.5\hspace{0.1em}/\hspace{0.1em}{\color{blue}\textbf{66.4}} & \textbf{22.8}\hspace{0.1em}/\hspace{0.1em}22.1\hspace{0.1em}/\hspace{0.1em}{\color{blue}21.7} & 62.1\hspace{0.1em}/\hspace{0.1em}62.4\hspace{0.1em}/\hspace{0.1em}{\color{blue}\textbf{67.7}} \\
    \lyft   & 77.5\hspace{0.1em}/\hspace{0.1em}82.2\hspace{0.1em}/\hspace{0.1em}{\color{blue}\textbf{82.3}} & 56.7\hspace{0.1em}/\hspace{0.1em}58.6\hspace{0.1em}/\hspace{0.1em}{\color{blue}\textbf{58.9}}& {\color{gray}78.3} & 24.1\hspace{0.1em}/\hspace{0.1em}23.6\hspace{0.1em}/\hspace{0.1em}{\color{blue}\textbf{26.3}} & 63.0\hspace{0.1em}/\hspace{0.1em}\textbf{69.1}\hspace{0.1em}/\hspace{0.1em}{\color{blue}68.7} \\
    \nusc   & 45.5\hspace{0.1em}/\hspace{0.1em}67.5\hspace{0.1em}/\hspace{0.1em}{\color{blue}\textbf{70.3}} & 19.5\hspace{0.1em}/\hspace{0.1em}36.1\hspace{0.1em}/\hspace{0.1em}{\color{blue}\textbf{42.7}} & 32.4\hspace{0.1em}/\hspace{0.1em}56.7\hspace{0.1em}/\hspace{0.1em}{\color{blue}\textbf{58.6}}& {\color{gray}36.8} & 42.6\hspace{0.1em}/\hspace{0.1em}60.2\hspace{0.1em}/\hspace{0.1em}{\color{blue}\textbf{61.1}} \\
    \waymo  & 74.1\hspace{0.1em}/\hspace{0.1em}76.6\hspace{0.1em}/\hspace{0.1em}{\color{blue}\textbf{77.3}} & 54.1\hspace{0.1em}/\hspace{0.1em}55.9\hspace{0.1em}/\hspace{0.1em}{\color{blue}\textbf{56.2}} & 66.0\hspace{0.1em}/\hspace{0.1em}68.1\hspace{0.1em}/\hspace{0.1em}{\color{blue}\textbf{68.7}} & 21.6\hspace{0.1em}/\hspace{0.1em}23.5\hspace{0.1em}/\hspace{0.1em}{\color{blue}\textbf{24.3}}& {\color{gray}71.3} \\\hline
\end{tabular}
\end{subtable}
\vspace{-15pt}
\end{table*}

\begin{table}[t]
	\tabcolsep 2 pt
	\renewcommand{\arraystretch}{1.05}
	\centering
    \fontsize{6.9}{8}\selectfont
	\caption{\textbf{UDA from \argo to \kitti.} We report \APBEV{}/ \AP of the \textbf{car} category at IoU $= 0.7$ and IoU $= 0.5$ across different depth range on the test set. NR stands for No-Retrain baseline, ST stands for Self-Training~\cite{kumar2020understanding}, SN stands for Statistical Normalization~\cite{yan2020domain}. Our method \emph{Dream} is marked in {\color{blue}blue}. We show the performance of in-domain model, \ie, the model trained and evaluated on \kitti, at the first row in {\color{gray}gray}. We also show results by directly applying online and offline (not feasible in real-time) tracking. Best viewed in color.}
	\label{tbl:k2a}
	\vspace{-.5em}
\begin{tabular}{=l|+c|+c|+c|+c|+c|+c}
    \multicolumn{1}{c|}{} & \multicolumn{3}{c|}{IoU 0.5} & \multicolumn{3}{c}{IoU 0.7} \\ \cline{2-7}
    \multicolumn{1}{c|}{\multirow{-2}{*}{Method}}
    & \rowstyle{\color{black}} 0-30    & 30-50    & 50-80   & 0-30    & 30-50    & 50-80   \\ \hline
    \rowstyle{\color{gray}} in-domain  & 90.0 / 89.9 & 81.0 / 79.9 & 40.4 / 36.3 & 89.0 / 78.0 & 70.3 / 51.5 & 26.6 / \phantom{0}9.8\\ \hline
    NR  & 89.4 / 88.8 & 71.0 / 65.2 & 18.0 / 13.9 & 72.6 / 47.8 & 35.8 / 14.6 & \phantom{0}4.9 / \phantom{0}3.0 \\
    NR + online  &89.3 / 88.5&71.9 / 66.1&18.3 / 14.0&72.3 / 47.5&38.2 / 14.9&\phantom{0}5.5 / \phantom{0}1.2\\
    \rowstyle{\color{gray}} NR + offline  & 89.3 / 88.8 & 72.7 / 67.7 & 18.9 / 15.0 & 74.5 / 49.6 & 43.5 / 18.1 & \phantom{0}7.3 / \phantom{0}1.8 \\\hline
    ST & \textbf{89.5} / \textbf{89.2} & 73.2 / 68.5 & 28.2 / 22.5 & 76.8 / 52.9 & 44.2 / 20.6 & 13.1 / \phantom{0}2.1 \\
    \rowstyle{\color{blue}}
    Dream   & 89.3 / \textbf{89.2} & \textbf{74.6} / \textbf{72.4} & \textbf{30.1} / \textbf{25.6} & \textbf{77.6} / \textbf{54.7} & \textbf{49.9} / \textbf{24.3} & \textbf{14.5} / \textbf{\phantom{0}3.4}\\ \hline\hline
    SN  & 89.3 / 88.2 & 69.6 / 65.4 & 14.6 / 13.3 & 83.8 / 59.1 & 53.5 / 27.2 & \phantom{0}9.3 / \phantom{0}3.6\\
    SN + online  &89.4 / 88.2&68.8 / 65.4&19.4 / 16.2&83.5 / 60.1&50.7 / 27.0&13.4 / \phantom{0}9.5\\
    \rowstyle{\color{gray}} SN + offline  & 89.4 / 88.2 & 68.9 / 66.1 & 21.8 / 19.3 & 83.3 / 59.2 & 50.9 / 26.9 & 13.8 / \phantom{0}9.8 \\\hline
    SN + ST  & \textbf{89.8} / 89.3 & 74.4 / 72.8 & 22.3 / 21.3 & \textbf{87.0} / 70.4 & 62.2 / 37.7 & 16.4 / \phantom{0}7.1\\
    \rowstyle{\color{blue}}
    SN + Dream  & \textbf{89.8} / \textbf{89.4} & \textbf{75.4} / \textbf{73.8} & \textbf{29.4} / \textbf{25.4} & \textbf{87.0} / \textbf{73.5} & \textbf{62.8} / \textbf{41.9} & \textbf{17.2} / \textbf{10.3}\\ \hline
\end{tabular}
\vspace{-15pt}
\end{table}
\textbf{Adaptation from \argo to \kitti.} We compare the UDA methods under \argo to \kitti in \autoref{tbl:k2a} and observe several trends: 
1) models experience a smaller domain gap when objects are closer (0-30 m vs 30-80 m); 2) though directly applying online tracking can improve the detection performance, models improve more after just self-training; 3) the offline tracking is used to provide extra pseudo-labels for fine-tuning, and interestingly, models fine-tuned from pseudo-labels can outperform pseudo-labels themselves; 4) \dreaming improves over ST and SN by a large margin, especially on IoU at 0.5;	
5) \dreaming has a large gain in AP for faraway objects, \eg, on range 50-80 m, compared to ST, it boosts the \APBEV on IoU at 0.5 from 28.2 to 30.1. 

\textbf{Adaptation among five datasets.} We further applied our methods to adaptation tasks among the five datasets. Due to limited space, we show the results of \APBEV and \AP on range 50-80~m and 0-80~m at IoU $=0.5$ in \autoref{tbl:5x5}. Our method consistently improves the adaptation performance on faraway ranges, while having mostly equal or better performance over the full range. We include detailed evaluation results across all ranges, at IoU $= 0.7$, and with SN in the supplementary material. We observe a consistent and clear trend as in \autoref{tbl:5x5}.

\begin{table}[!t]
    \vspace{5pt} 
	\tabcolsep 1.5pt
	\fontsize{7}{8}\selectfont
	\renewcommand{\arraystretch}{1.05}
	\centering
	\caption{\textbf{UDA from \kitti(city, campus) to \kitti(road, residential)}. Naming is as in \autoref{tbl:k2a}.
	\label{tbl:kitticc2kittirr}}
	\vspace{-.5em}
    \begin{tabular}{=l|+c|+c|+c|+c|+c|+c}
    \multicolumn{1}{c|}{} & \multicolumn{3}{c|}{IoU 0.5} & \multicolumn{3}{c}{IoU 0.7} \\ \cline{2-7}
    \multicolumn{1}{c|}{\multirow{-2}{*}{Method}}            &\color{black} 0-30    & \color{black}30-50    & \color{black}50-80   & \color{black}0-30    & \color{black}30-50    & \color{black}50-80   \\ \hline
   NR & 89.4 / 89.4 & 79.7 / 77.9 & 36.2 / 30.9 & 88.3 / 77.9 & 66.0 / 47.5 & 19.2 / \phantom{0}6.2  \\ 
    ST      &89.4 / 89.4 & 78.0 / 76.7 & 35.9 / 31.2 & 88.2 / 77.8 & 66.3 / 48.3 & 19.0 / \phantom{0}7.7 \\
    \rowstyle{\color{blue}} Dream & \textbf{89.6 / 89.6} & \textbf{80.1 / 78.9} & \textbf{41.4 / 35.3} & \textbf{88.5 / 78.2} & \textbf{68.0 / 49.6} & \textbf{23.2 / 12.7} \\\hline
    \end{tabular}
    \vspace{-10pt}
\end{table}
\textbf{Adaptation between different locations inside the same dataset.} Different datasets not only come from different locations but also use different sensor configurations. To isolate the effects of the former (which is our motivating application), in  \autoref{tbl:kitticc2kittirr} we evaluate our method's performance for domain adaptation within the \kitti dataset. 
In this case, the source and target domain are all scenes from \kitti, while the source is composed of city and campus scenes (38 sequences, 9,556 frames, 3,420 labeled frames) and the target consists of residential and road scenes (23 sequences, 10,448 frames, 3619 labeled frames). 
Our method consistently outperforms no fine-tuning and ST, especially on 30-80 m range.

\begin{table}[!bt]
	\tabcolsep 1.5pt
	\fontsize{6.7}{8}\selectfont
	\renewcommand{\arraystretch}{1.05}
	\centering
	\caption{\textbf{Ablation study of UDA from \argo to \kitti.} We report \APBEV / \AP of the \textbf{car} category at IoU $= 0.5$ and IoU $= 0.7$ across different depth range, using \method{PointRCNN} model. Naming is as in \autoref{tbl:k2a}. S stands for smoothing, R for resizing, I for interpolation and E for extrapolation. \label{tbl:k2a_ablation}}
	\vspace{-.5em}
    \begin{tabular}{=l|+c|+c|+c|+c|+c|+c}
    \multicolumn{1}{c|}{} & \multicolumn{3}{c|}{IoU 0.5} & \multicolumn{3}{c}{IoU 0.7} \\ \cline{2-7}
    \multicolumn{1}{c|}{\multirow{-2}{*}{Method}}            & \color{black}0-30    & \color{black}30-50    & \color{black}50-80   & \color{black}0-30    & \color{black}30-50    & \color{black}50-80   \\ \hline
    NR           & 89.4 / 88.8 & 71.0 / 65.2 & 18.0 / 13.9 & 72.6 / 47.8 & 35.8 / 14.6 & \phantom{0}4.9 / \phantom{0}3.0 \\ 
    ST                 & 89.5 / 89.2 & 73.2 / 68.5 & 28.2 / 22.5 & 76.8 / 52.9 & 44.2 / 20.6 & 13.1 / \phantom{0}2.1\\
    ST + S      & 89.5 / 89.3 & 73.7 / 68.7 & 28.3 / 22.3 & 76.8 / 53.4 & 45.2 / 21.5 & 12.9 / \textbf{\phantom{0}4.7} \\
    ST + S + R &\textbf{89.5 / 89.3} & 74.0 / 71.6 & 28.3 / 23.9& \textbf{78.0 / 55.2} & \textbf{50.8} / 23.9 & 10.3 / \phantom{0}2.7\\
    ST + S + R + I &89.3 / 89.1&73.9 / 71.6&28.1 / 23.3&77.6 / 54.5&50.4 / 24.0&11.2 / \phantom{0}3.4\\
    ST + S + R + I + E &89.4 / 89.2&\textbf{74.9 / 72.5}&\textbf{31.0 / 25.7}&77.8 / 55.1&50.4 / \textbf{24.1}&\textbf{14.3} / \phantom{0}3.3\\ \hline
    \end{tabular}
    \vspace{-20pt}
\end{table}
\textbf{Ablation Study.} We show ablation results in  \autoref{tbl:k2a_ablation}. Here we fine-tune models using ST and adding smoothing (S), resizing (R), interpolation (I) and extrapolation (E) to the pseudo-label generation. It can be observed that ST alone already boosts performance considerably. Through selecting high confidence detections, smoothing and adjusting the object size we ensure that the pseudo-labels provided are mostly \emph{correct}. But just these do not address the second criteria for desired pseudo-labels: \emph{high coverage}. We observe noticeable boosts when interpolation and extrapolations are added, specially for far away objects. This is due to extrapolations and interpolations recovering pseudo-labels for low confidence or missed detections for distant vehicles.

\begin{table}[t]
	\tabcolsep 1.5pt
	\renewcommand{\arraystretch}{1.05}
	\centering
	\fontsize{6.9}{8}\selectfont
	\caption{\textbf{UDA from \argo to \kitti using \PIXOR.} We report \APBEV of the \textbf{car} category at  IoU $= 0.5$ and $0.7$. Naming is as in \autoref{tbl:k2a}. \label{tbl:k2a_pixor}}
	\vspace{-.5em}
    \begin{tabular}{=l|+C{23pt}|+C{23pt}|+C{23pt}|+C{23pt}|+C{23pt}|c}
    \multicolumn{1}{c|}{} & \multicolumn{3}{c|}{IoU 0.5} & \multicolumn{3}{c}{IoU 0.7} \\ \cline{2-7}
    \multicolumn{1}{c|}{\multirow{-2}{*}{Method}}            \rowstyle{\color{black}} & 0-30    & 30-50    & 50-80   & 0-30    & 30-50    & 50-80   \\ \hline
    \rowstyle{\color{gray}} in-domain        & 88.7 & 62.6 & 21.4 & 79.6 & 49.9 & 10.0 \\ \hline
    NR           &85.7 & 57.2 & 12.9 & 54.2 & 23.6 & \phantom{0}4.7\\ 
    ST                  &86.0 & 56.3 & 12.2 & 55.3 & 24.6 & \phantom{0}2.5\\
    \rowstyle{\color{blue}}
    Dream       &\textbf{87.1} & \textbf{61.1} & \textbf{20.2} & \textbf{58.0} & \textbf{28.1} & \textbf{\color{blue} \phantom{0}4.8}\\ \hline
    SN       &86.7 & 58.7 & 15.1 & 76.2 & 38.7 & \phantom{0}5.1\\
    SN + ST             &\textbf{87.4} & 58.9	 & 12.4 & \textbf{78.0} & 42.4 & \phantom{0}3.8\\
    \rowstyle{\color{blue}}
    SN + Dream  &\textbf{87.4} & \textbf{64.2} & \textbf{22.2} & 77.9 & \textbf{42.5} & \textbf{ \color{blue} \phantom{0}4.5}\\ \hline
    \end{tabular}
    \vspace{-20pt}
\end{table}

\textbf{Adaptation results using PIXOR.} 
To show the generality of our approach, we further apply it to another detector \PIXOR~\cite{yang2018pixor} from \argo to \kitti in \autoref{tbl:k2a_pixor}. Dreaming improves the accuracy at farther ranges (30-80 m) while maintaining the accuracy at close range (0-30 m). Interestingly, at IoU 0.5 in the 30-80 m ranges, we are able to surpass the in-domain performance, which uses models trained only in the target domain with the ground-truth labels. This results showcases the power of unsupervised domain adaptation (UDA): with a suitably designed algorithm, UDA that leverages both the source and target domain could outperform models trained only in a single domain.

\textbf{Others.} We show more results and qualitative visualizations in the supplementary material.

\vspace{-0.2cm}
\section{Conclusion and Discussion}

In this paper, we have introduced a novel method towards closing the gap between  source and target in unsupervised domain adaptation for LiDAR-based 3D object detection. Our approach is based on self-training, while leveraging vehicle dynamics and offline analysis to generate pseudo-labels. Importantly, we can generate high quality pseudo-labels even for difficult cases (i.e. far-away objects), which the detector tends to miss before adaptation. Fine-tuning on these pseudo-labels improves  detection performance drastically in the target domain. It is hard to conceive an autonomous vehicle manufacturer that could collect, label, and update data for every consumer environment, meeting the requirements to allow self-driving cars to operate everywhere freely and safely. By significantly reducing the adaptation gap between domains, our approach takes a significant step towards making this vision a reality nevertheless.

\section*{Acknowledgment}
{This research is supported by grants from the National Science Foundation NSF (III-1618134, III-1526012, IIS-1149882, IIS-1724282, TRIPODS-1740822, IIS-2107077, OAC-2118240, OAC-2112606), the Office of Naval Research DOD (N00014-17-1-2175), the Bill and Melinda Gates Foundation, and the Cornell Center for Materials Research with funding from the NSF MRSEC program (DMR-1719875).}

\bibliographystyle{IEEEtran}
\bibliography{main}

\appendix

\subsection{Hyper-parameters}
In performing the forward pass of the Extended Kalman Filter (EKF) to calculate the a-posteriori state and state error covariance estimates (\ie, $\bar{\mathbf{x}}_{k|k}$ $ P_{k|k}$) and the a-priori state and state error covariance (\ie,  $\bar{\mathbf{x}}_{k+1|k}$  $P_{k+1|k}$) in \autoref{ssec:PL_T}, a process noise covariance matrix ($Q$), measurement noise covariance matrix ($R$), initial state estimate ($\mathbf{\bar{x}}_{0}$), and  initial state error covariance matrix ($P_0$) are necessary. The measurement noise matrix is established through measurement variance, and was obtained by observing the variance of position ($x,y$), orientation ($\theta$), length ($l$) and width ($w$) errors from testing detector in \argo to \kitti setting. A larger measurement noise matrix is used in the EKF for extrapolation as larger variance was observed with far-away detections. The process noise matrix should represent the magnitudes of dynamical noise that the system might experience. In the model in \cite{8843260}, which is constant velocity and heading, there are seven noise parameters: $e_\theta, e_s, e_{v_x}, e_{v_y},  e_{\omega_z},e_l,e_w$ which correspond to the diagonal of the $Q$ matrix. The variables $e_\theta,e_s,e_l,e_w$ are modeled as zero mean, mutually uncorrelated, Gaussian, and white process noise. Intuitively, $e_\theta,e_s$ represent the uncertainty associated with the orientation and speed of vehicles, especially given that the model assumes constant speed straight line motion. The noise associated with the orientation and speed of the target vehicles are the largest and of most importance. The time derivative of the objects length and width are zero, where  $e_l,e_w$ are small tuning parameters which control the response of the filter. While $e_{v_x}$, $e_{v_y}$,  $e_{\omega_z}$ are noises associated with the pose of ego-vehicle and small values were chosen given that a high accuracy in ego-vehicle pose is expected across the datasets. Finally, the EKF necessitates an initialization for the state and state error covariance estimates. The initial detection was used for state initialization, and relatively large conservative uncertainties were used for state error covariance.

\label{ssec:hyper_parameters}
\begin{itemize}
     \item Extended Kalman Filter (EKF) and Global Nearest Neighbor (GNN) parameters for tracking and data association (cf. \autoref{ssec:PL_T}):
     \begin{enumerate}
    \item Measurement noise covariance matrix:\\
	\hspace{20pt} $R=\text{diag}(0.1 \text{m}^2,  0.1 \text{m}^2, 0.015 \text{rad}^2,  0.07  \text{m}^2,  0.04  \text{m}^2)$
    \item Process noise covariance matrix:\\
     \hspace{20pt}  $Q=\text{diag}(0.1218 \frac{\text{rad}}{\text{s}}^2,  1 \frac{\text{m}}{\text{s}^2}^2, 0.00545 \frac{\text{rad}}{\text{s}}^2,  0.00545 \frac{\text{m}}{\text{s}}^2,\\  0.00307 \frac{\text{rad}}{\text{s}}^2, 0.01\frac{\text{m}}{\text{s}}^2, 0.01\frac{\text{m}}{\text{s}}^2) $
    \item State error covariance matrix initialization: \\
    \hspace{20pt}  $P_0=\text{diag}(2 \text{m}^2,2 \text{m}^2,0.1 \text{rad}^2, 5 \frac{\text{m}}{\text{s}}^2,  0.5 \text{m}^2, 0.3 2 \text{m}^2)$
     \item The initial state estimate, $\mathbf{\bar{x}}_{0}$, is set to be the first detection values.

     \item The data association threshold (in \textbf{BEV IoU}) in GNN is set to be 0.3.

     \item The fraction of distance between the vehicle center and back-axle is $\frac{1}{4}l$, as in \cite{8843260}.

	\end{enumerate}
    \item EKF parameters for extrapolation:
    \begin{enumerate}
    	\item Measurement noise covariance matrix:\\
    	\hspace{20pt} $R=\text{diag}(0.5 \text{m}^2,  0.5 \text{m}^2, 0.06 \text{rad}^2,  0.07  \text{m}^2,  0.04  \text{m}^2)$
    \end{enumerate}
\end{itemize}

We set $c_{\mathrm{min-hits}} = 3$ and $c_{\mathrm{max-age}} = 3$. When doing extrapolation, we use $-25$ and $-3$ as thresholds for \method{PointRCNN} and \PIXOR models respectively.

\subsection{Details on Datasets}
\label{ssec:dataset}

We split each dataset into two parts, a training set and a test set. When a dataset is used as the source, we train the detector using its (ground truth) training labeled frames. When a dataset is used as the target, we adapt the detector using its training sequences without revealing the ground truth labels. We evaluate the adapted models on the test set.
We provide detailed properties of the five autonomous driving datasets and the way we split the data as follows.

\textbf{\kitti.} The \kitti object detection benchmark~\cite{geiger2013vision,geiger2012we} contains 7,481 scenes for training and 7,518 scenes for testing.
All the scenes are pictured around Karlsruhe, Germany in clear weather and day time.
For each scene, \kitti provides 64-beam Velodyne LiDAR point cloud and stereo images. The training set is further separated into 3,712 training and 3,769 validation scenes as suggested by \cite{chen20153d}.
The training scenes are sampled from 96 data sequences, which have no overlap with the sequences where validation scenes are sampled.
These training sequences are collected in 10 Hz, resulting in 13,596 frames. We extract the sequences from the raw \kitti data as our adaptation data. We use such data splits in all experiments related to \kitti, except for \textbf{adaptation between different locations inside the same dataset} (cf. \autoref{tbl:kitticc2kittirr} of the main paper). For adaptation between different locations inside the \kitti dataset, we split the sequences in training set by their categories: \emph{city/campus} as the source (38 sequences, 9,556 frames, 3,420 labeled frames) and \emph{residential/road} as the target (23 sequences, 10,448 frames, 3619 labeled frames). In \autoref{tbl:kitticc2kittirr}, all models are pre-trained in city/campus data. Due to limited data, different from what we do UDA among five autonomous driving datasets, \emph{we perform adaptation and final evaluation on the full target data.} Note that in this case, the detectors still do not have access to ground truth labels during UDA.

\textbf{\argo.} The \argo dataset~\cite{argoverse} is collected around Miami and Pittsburgh, USA in multiple weathers and during different times of the day.
For each scene (timestamp), \argo provides a 64-beam LiDAR point cloud captured by stacking two 32-beam Velodyne LiDAR vertically.
We extracted synchronized images (from front camera) and corresponding point clouds from the original \argo dataset. We follow the official split on training and validation sets, which contain 13,122 scenes and 5,014 scenes respectively. We use sequences in the training set (without using the ground truth labels) as our adaptation data.

\textbf{\lyft.} The \lyft Level 5 dataset collects 18,634 scenes around Palo Auto, USA in clear weather and during day time. For each scene, \lyft provides the ground-truth bounding box labels and point cloud captured by a 40 (or 64)-beam roof LiDAR and two 40-beam bumper LiDAR sensors. We follow \cite{yan2020domain} and separate the dataset by sequences, resulting in 12,599 frames for training (100 sequences), 3,024  validation frames (24 sequences) and 3,011 frames for testing (24 sequences). The sequences in 5 Hz and we use training sequences without labels as adaptation data.

\textbf{\nusc.} The \nusc dataset~\cite{nuscenes2019} collects scenes around Boston, USA and Singapore in multiple weather conditions and during different times of the day. For each scene, \nusc provides a point cloud captured by a 32-beam roof LiDAR. We use the data collected in Boston, and sampled 312 sequences for training and 78 sequences for validation. We use 10 Hz sensor data without labels as our adaptation data (61,121 frames in total), and evaluate the model on 2 Hz labeled data in validation set (3,133 frames). Note that after generating pseudo-labels, we sub-sample adaptation data using 2 Hz into 12,562 frames.

\textbf{\waymo.} The \waymo open dataset~\cite{waymo_open_dataset} is mostly collected around San Francisco, Phoenix, and Mountain View in multiple weather conditions and at multiple times of the day. It provides point clouds captured in 10 Hz by five LiDAR sensors (one on roof, four on side) and images from five cameras. We randomly sample 60 tracks (11,886 frames) captured in San Francisco as our adaptation data, and another 100 tracks (19,828 frames) in San Francisco as our validation data.

\subsection{Qualitative Results}
\label{ssec:qualitative}
\begin{figure*}[t]
	\centering
	\begin{subfigure}{\textwidth}
		\centering
		\includegraphics[width=\textwidth]{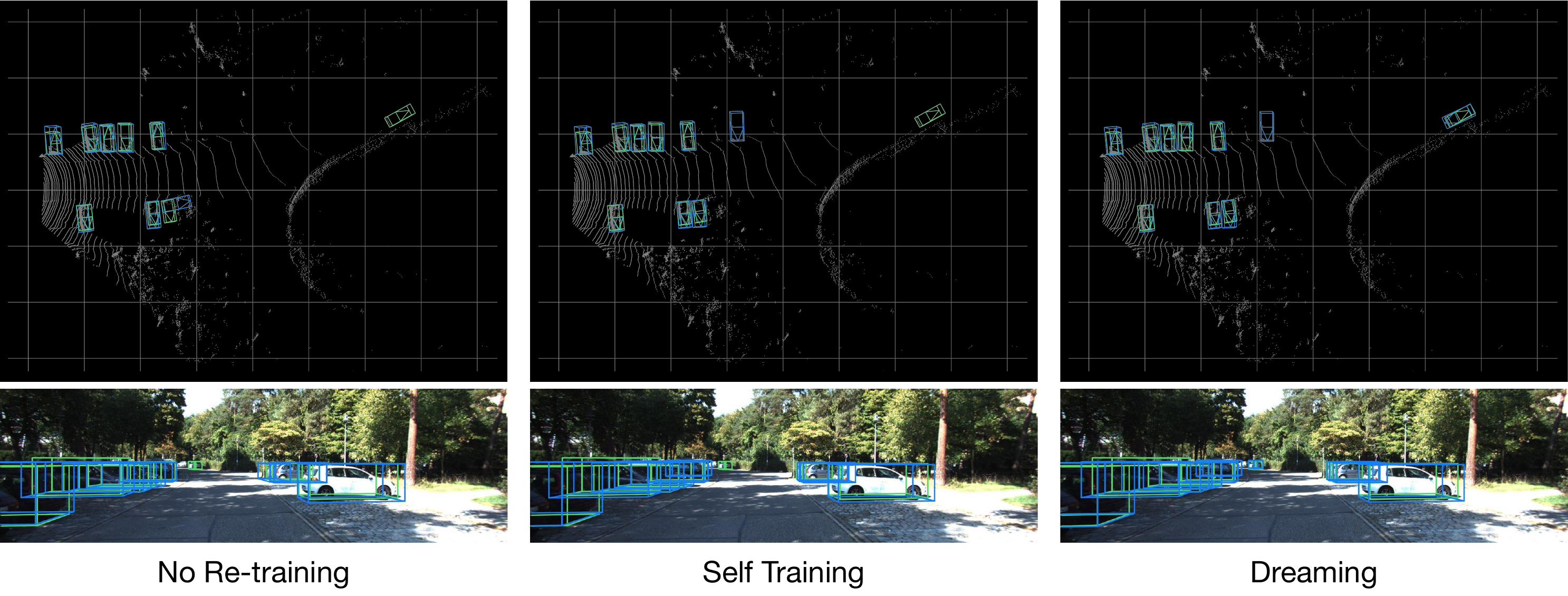}
	\end{subfigure}
	\vfill

	\begin{subfigure}{\textwidth}
		\centering
		\includegraphics[width=\textwidth]{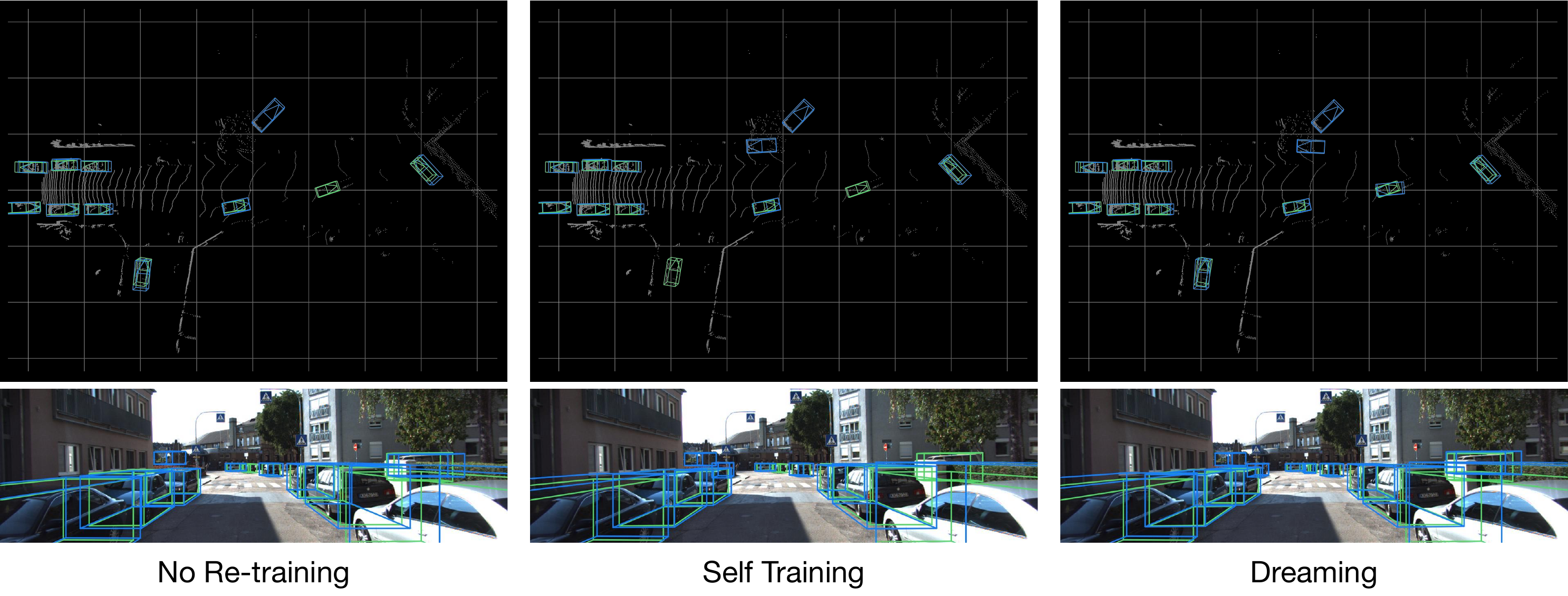}
	\end{subfigure}
	\vfill

	\begin{subfigure}{\textwidth}
		\centering
		\includegraphics[width=\textwidth]{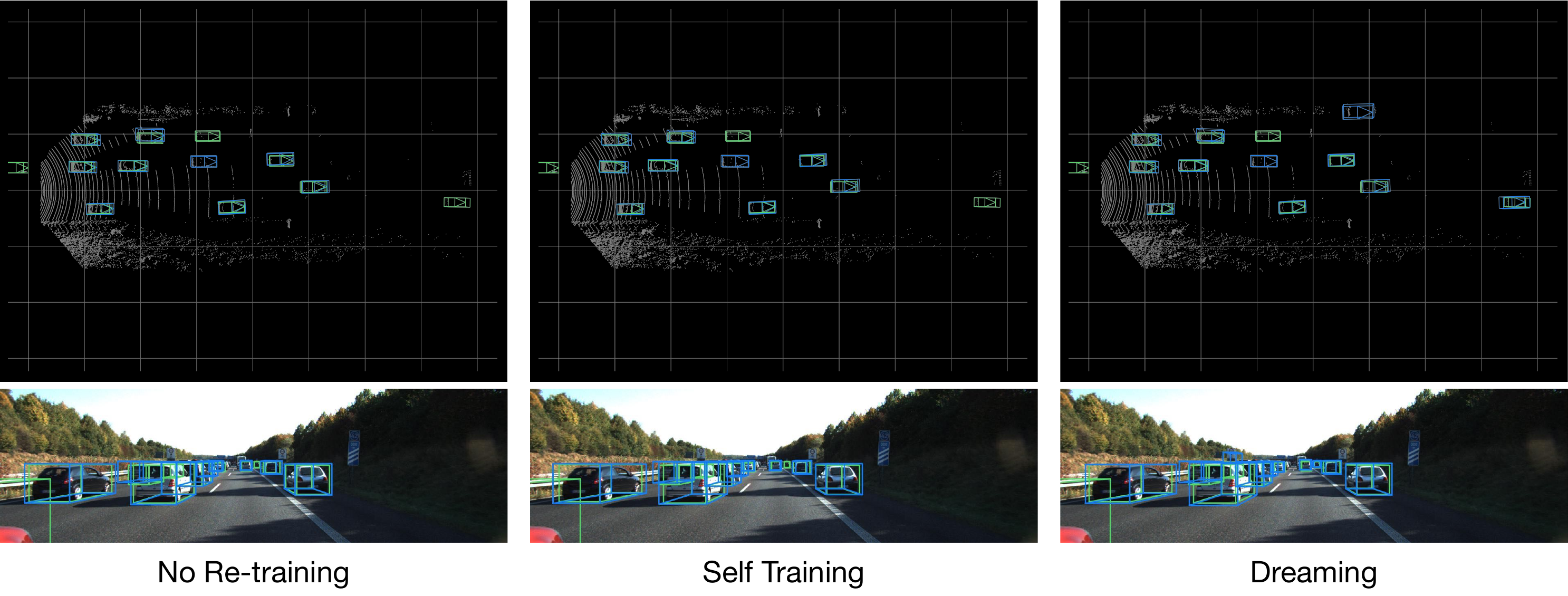}
	\end{subfigure}
	\caption{\textbf{Qualitative Results.} We compare the detection results on several scenes from the \kitti validation set by the \method{PointRCNN} detectors that are trained on 1) \argo dataset (No Re-training), 2) \argo dataset and fine-tuned using self-training on \kitti (Self Training), and  3) \argo dataset and fine-tuned using dreaming on \kitti (Dreaming). We visualize them from both frontal-view images and bird's-eye view point maps. Ground-truth boxes are in green and detected bounding boxes are in blue. The ego vehicle is on the left side of the BEV map and looking to the right. One floor square is 10m$\times$10m. Best viewed in color. Zoom in for details. \label{fig:qualitative_argo2kitti}}
\end{figure*}

\begin{figure*}[t]
	\centering
	\begin{subfigure}{\textwidth}
		\centering
		\includegraphics[width=\textwidth]{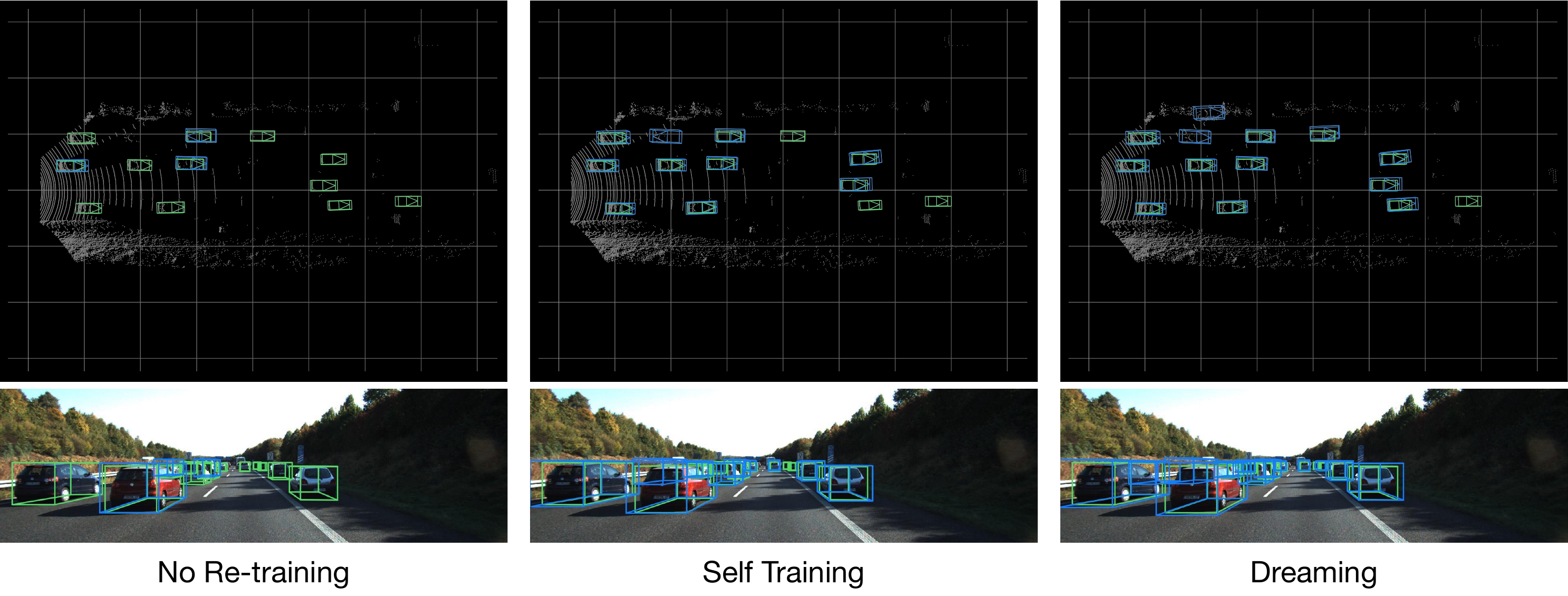}
	\end{subfigure}
	\vfill

	\begin{subfigure}{\textwidth}
		\centering
		\includegraphics[width=\textwidth]{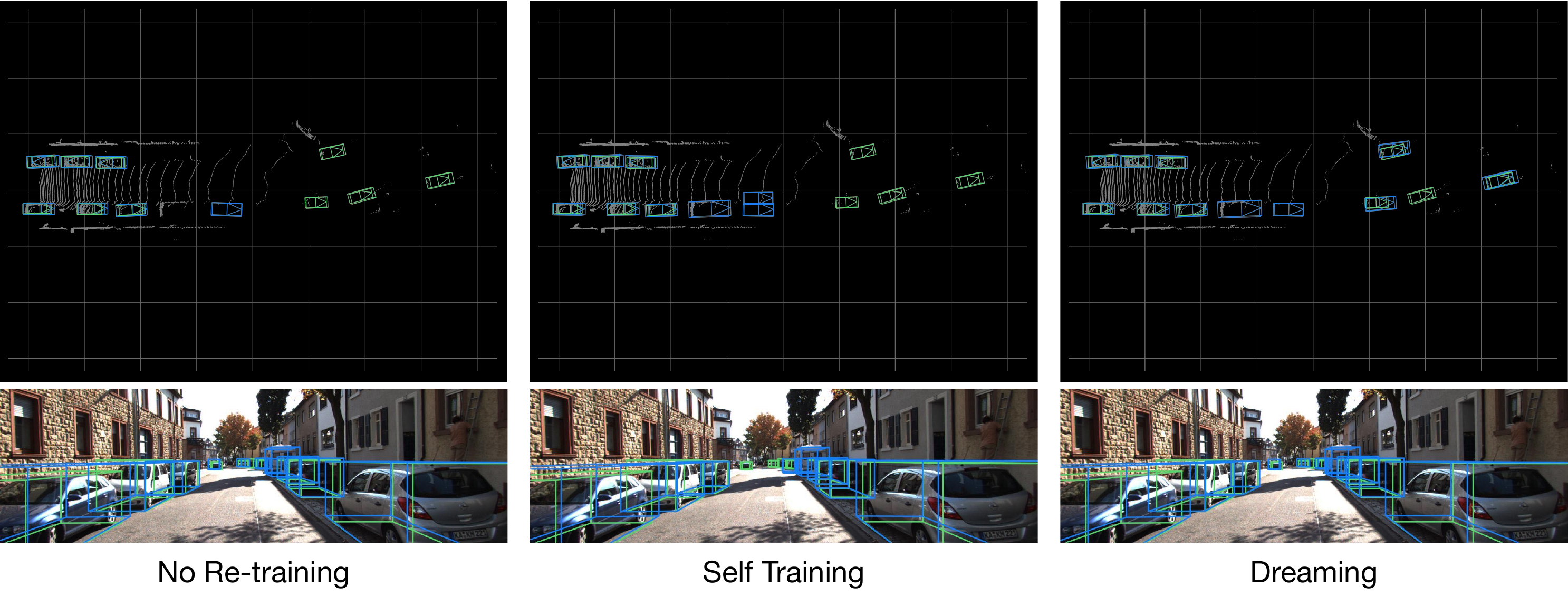}
	\end{subfigure}
	\vfill

	\begin{subfigure}{\textwidth}
		\centering
		\includegraphics[width=\textwidth]{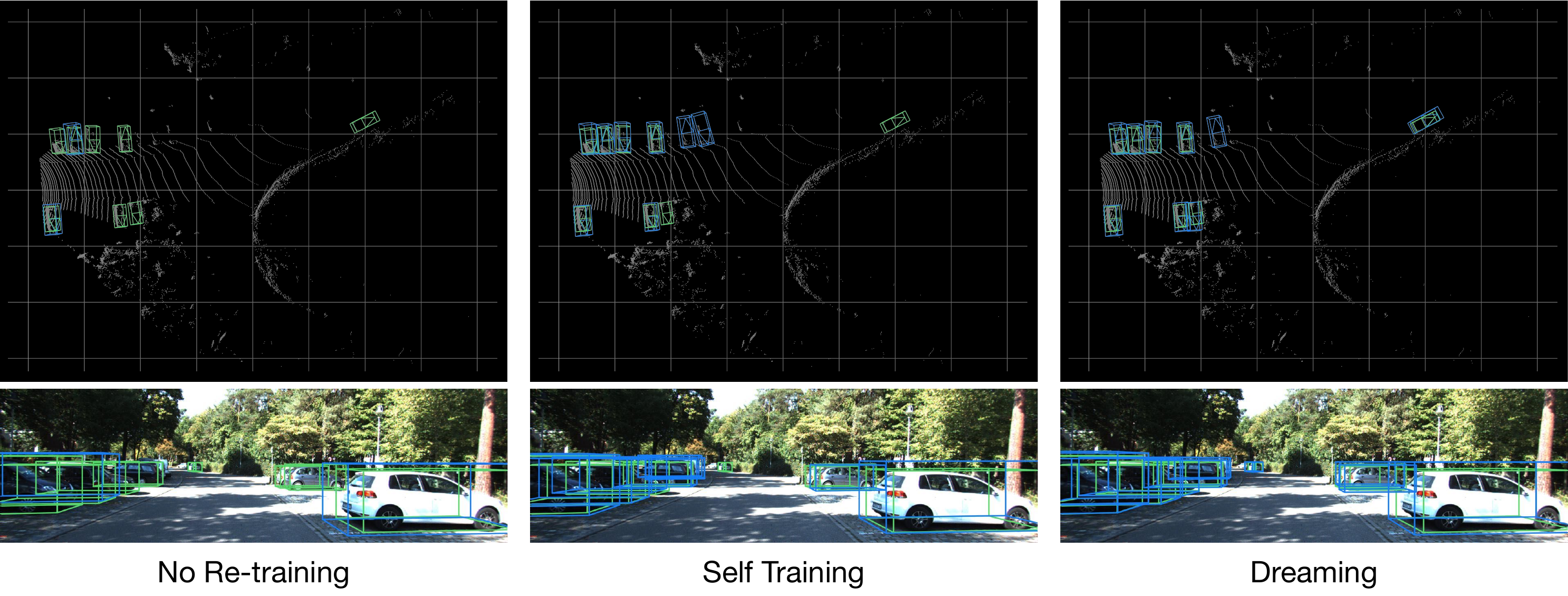}
	\end{subfigure}
	\caption{\textbf{Qualitative Results.} The setups are the same as those in \autoref{fig:qualitative_argo2kitti}, but the models are pre-trained in \nusc dataset and tested on \kitti dataset. \label{fig:qualitative_nusc2kitti}}
\end{figure*}

\begin{figure*}[t]
	\centering
	\begin{subfigure}{.9\textwidth}
		\centering
		\includegraphics[width=\textwidth]{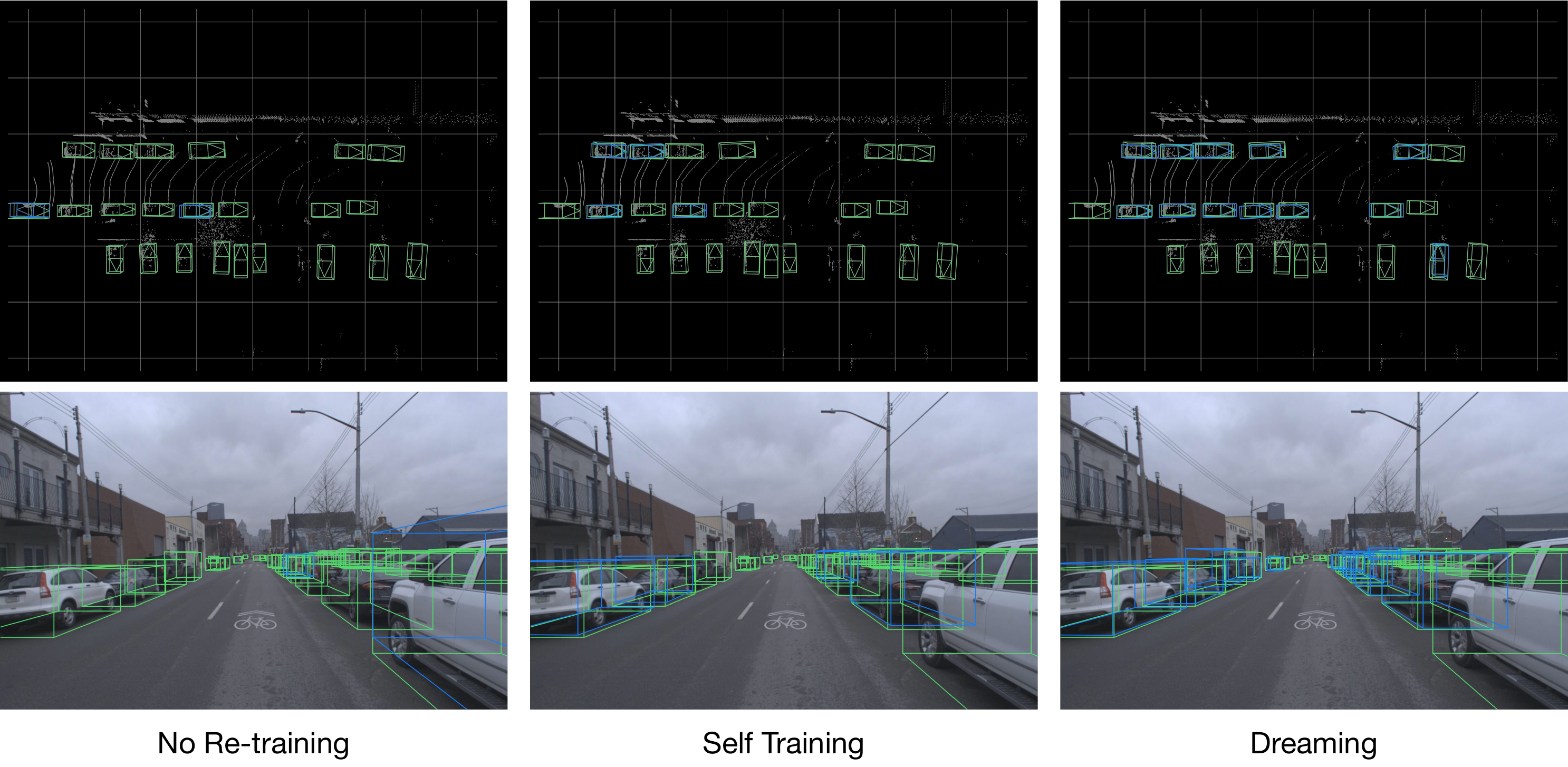}
	\end{subfigure}
	\vfill

	\begin{subfigure}{.9\textwidth}
		\centering
		\includegraphics[width=\textwidth]{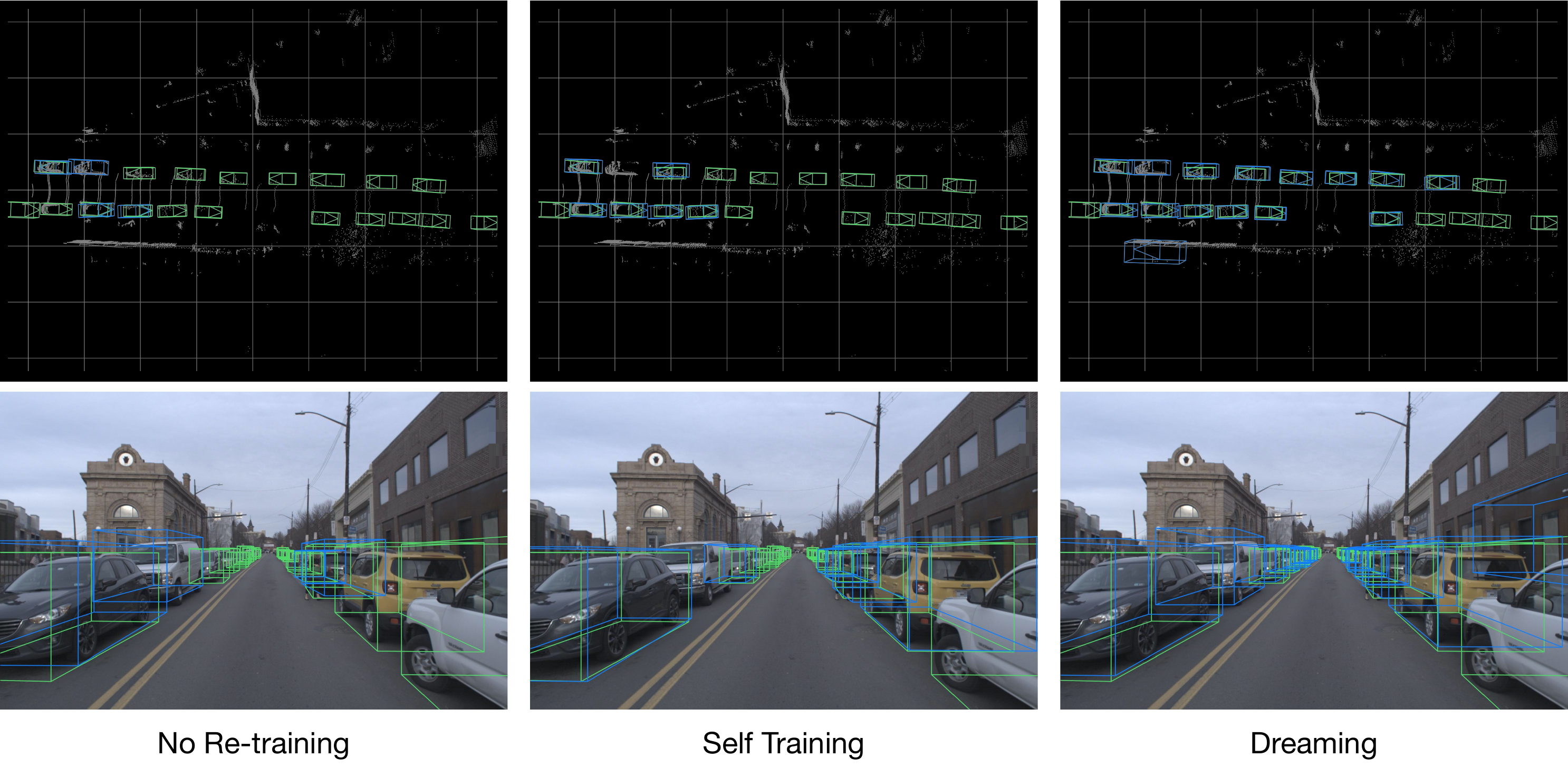}
	\end{subfigure}
	\vfill

	\begin{subfigure}{.9\textwidth}
		\centering
		\includegraphics[width=\textwidth]{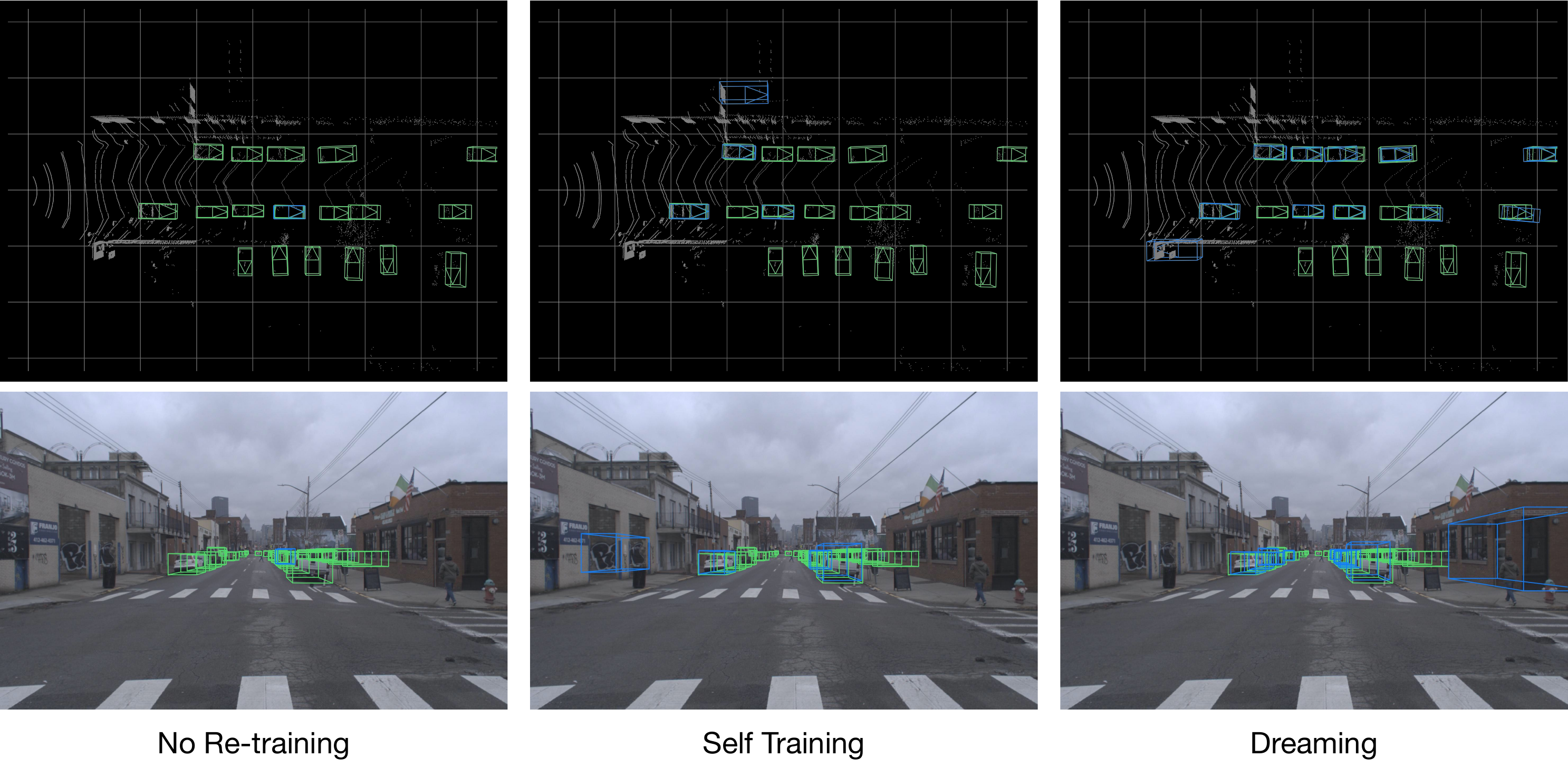}
	\end{subfigure}
	\caption{\textbf{Qualitative Results.} The setups are the same as those in \autoref{fig:qualitative_argo2kitti}, but the models are pre-trained in \nusc dataset and tested on \argo dataset. \label{fig:qualitative_nusc2argo}}
\end{figure*}

In \autoref{fig:qualitative_argo2kitti}, \autoref{fig:qualitative_nusc2kitti} and \autoref{fig:qualitative_nusc2argo}, we compare qualitatively the detection results from models trained with different adaptation strategies. We select (\argo, \kitti), (\nusc, \kitti) and (\nusc, \argo) as (\emph{source}, \emph{target}) example pairs in qualitative visualization. It can be seen that models without retraining tend to miss faraway objects. Models with self-training are able to detect some of these objects. Models with dreaming can detect more faraway objects. Self-training and dreaming both exhibit some more false positive detection.

\subsection{Adaptation among Five Datasets}
\label{ssec:5x5}
In \autoref{suppl-tbl:5x5}, we present UDA results on all possible (source, target) pairs among the five autonomous driving datasets (20 pairs in total). On each pair, we show results with and without statistical normalization \cite{yan2020domain}. As in \autoref{tbl:k2a}, we report \APBEV / \AP of the \textbf{car} category at IoU $= 0.7$ and IoU $= 0.5$ across different depth range, using the \method{PointRCNN} detector. It can be seen that under these 20 UDA scenarios, our method consistently improves the adaptation performance on faraway ranges, while having mostly equal or better performance on close-by ranges.

\begin{table*}[ht]
\tabcolsep 4pt
\renewcommand{\arraystretch}{1.05}
\centering
\caption{\textbf{Unsupervised domain adaptation among five autonomous driving datasets.} Naming is as that in \autoref{tbl:k2a} of the main paper. \label{suppl-tbl:5x5}}
\begin{subtable}{\textwidth}
    \caption{\kitti to \argo}
    \centering
    \begin{tabular}{=l|+c|+c|+c|+c|+c|+c}
        \multicolumn{1}{c|}{} & \multicolumn{3}{c|}{IoU 0.5} & \multicolumn{3}{c}{IoU 0.7} \\ \cline{2-7}
        \multicolumn{1}{c|}{\multirow{-2}{*}{\diagbox{Method}{Range(m)}}}            & \rowstyle{\color{black}} 0-30    & 30-50    & 50-80   & 0-30    & 30-50    & 50-80   \\ \hline
        \rowstyle{\color{gray}} in-domain  & 88.6 / 85.7 & 68.5 / 66.4 & 37.8 / 30.0 & 76.5 / 53.2 & 56.6 / 30.4 & 20.2 / 10.1\\ \hline
        no retraining  & 79.4 / 76.5 & 51.9 / 44.6 & 20.1 / 15.4 & 53.1 / 32.9 & 29.3 / 12.3 & \phantom{0}5.9 / \phantom{0}3.0 \\
        ST  & 85.2 / 77.6 & 56.2 / 52.2 & 26.9 / 19.5 & 61.3 / \textbf{38.8} & 34.9 / 17.4 & 10.9 / \textbf{\phantom{0}9.1} \\
        \rowstyle{\color{blue}}
        Dreaming   & \textbf{85.6} / \textbf{77.8} & \textbf{61.3} / \textbf{54.1} & \textbf{28.4} / \textbf{20.8} & \textbf{63.3} / 38.6 & \textbf{41.6} / \textbf{20.2} & \textbf{12.4} / \phantom{0}4.5\\ \hline
        SN only  & 79.3 / 77.6 & 54.7 / 52.0 & 27.5 / 21.2 & 66.2 / 48.5 & 43.8 / 21.2 & 16.7 / \phantom{0}9.1\\
        SN + ST  & 84.9 / 78.0 & 56.7 / 54.7 & 29.0 / 22.5 & 73.0 / \textbf{50.8} & 46.4 / \textbf{22.3} & 17.6 / \phantom{0}9.1\\
        \rowstyle{\color{blue}}
        SN + Dreaming  & \textbf{85.1} / \textbf{78.2} & \textbf{62.3} / \textbf{55.7} & \textbf{33.6} / \textbf{26.5} & \textbf{73.1} / 50.3 & \textbf{50.6} / 20.6 & \textbf{18.5} / \textbf{\phantom{0}9.9}\\ \hline
    \end{tabular}
\end{subtable}

\begin{subtable}{\textwidth}
    \caption{\kitti to \lyft}
    \centering
    \begin{tabular}{=l|+c|+c|+c|+c|+c|+c}
        \multicolumn{1}{c|}{} & \multicolumn{3}{c|}{IoU 0.5} & \multicolumn{3}{c}{IoU 0.7} \\ \cline{2-7}
        \multicolumn{1}{c|}{\multirow{-2}{*}{\diagbox{Method}{Range(m)}}}            & \rowstyle{\color{black}} 0-30    & 30-50    & 50-80   & 0-30    & 30-50    & 50-80   \\ \hline
        \rowstyle{\color{gray}} in-domain  & 89.2 / 88.9 & 78.4 / 77.9 & 67.2 / 65.5 & 88.4 / 79.0 & 74.8 / 54.6 & 54.7 / 27.3\\ \hline
        no retraining  & 81.0 / \textbf{80.8} & 73.6 / 66.8 & 49.6 / 40.0 & 69.7 / 50.4 & 48.0 / 24.5 & 25.4 / \phantom{0}5.7 \\
        ST  & \textbf{85.9} / 80.7 & 74.1 / 67.9 & 56.3 / 46.3 & 75.2 / 53.7 & 53.8 / 25.3 & 28.9 / \phantom{0}7.6 \\
        \rowstyle{\color{blue}}
        Dreaming   & \textbf{85.9} / 80.7 & \textbf{74.8} / \textbf{68.6} & \textbf{56.4} / \textbf{47.7} & \textbf{76.6} / \textbf{54.2} & \textbf{56.3} / \textbf{31.3} & \textbf{34.4} / \textbf{10.2}\\ \hline
        SN only  & \textbf{81.0} / \textbf{80.9} & 72.6 / 67.2 & 55.0 / 47.5 & \textbf{78.7} / \textbf{67.3} & 64.9 / 45.1 & 43.4 / 18.0\\
        SN + ST  & 80.6 / 80.5 & 73.5 / 72.5 & 56.0 / 48.8 & 78.1 / 65.9 & 66.6 / 45.5 & 45.9 / 18.7\\
        \rowstyle{\color{blue}}
        SN + Dreaming  & 80.5 / 80.3 & \textbf{74.3} / \textbf{72.9} & \textbf{56.6} / \textbf{49.7} & 78.1 / 65.9 & \textbf{67.0} / \textbf{48.5} & \textbf{46.7} / \textbf{22.0}\\ \hline
    \end{tabular}
\end{subtable}

\begin{subtable}{\textwidth}
    \caption{\kitti to \nusc}
    \centering
    \begin{tabular}{=l|+c|+c|+c|+c|+c|+c}
        \multicolumn{1}{c|}{} & \multicolumn{3}{c|}{IoU 0.5} & \multicolumn{3}{c}{IoU 0.7} \\ \cline{2-7}
        \multicolumn{1}{c|}{\multirow{-2}{*}{\diagbox{Method}{Range(m)}}}            & \rowstyle{\color{black}} 0-30    & 30-50    & 50-80   & 0-30    & 30-50    & 50-80   \\ \hline
        \rowstyle{\color{gray}} in-domain  & 65.6 / 64.5 & 18.4 / 17.2 & \phantom{0}3.5 / \phantom{0}2.8 & 57.8 / 37.0 & 15.8 / \phantom{0}5.3 & \phantom{0}2.6 / \phantom{0}0.8\\ \hline
        no retraining  & 48.6 / 40.7 & 11.0 / \phantom{0}4.5 & \phantom{0}1.4 / \textbf{\phantom{0}0.9} & 33.3 / 13.4 & \phantom{0}4.5 / \phantom{0}0.7 & \phantom{0}0.3 / \textbf{\phantom{0}0.0} \\
        ST  & 52.2 / 43.9 & 11.8 / \phantom{0}9.1 & \textbf{\phantom{0}9.1} / \phantom{0}0.4 & 33.2 / \phantom{0}8.6 & \phantom{0}9.1 / \textbf{\phantom{0}4.5} & \phantom{0}3.0 / \textbf{\phantom{0}0.0} \\
        \rowstyle{\color{blue}}
        Dreaming   & \textbf{53.6} / \textbf{45.8} & \textbf{13.8} / \textbf{\phantom{0}9.7} & \phantom{0}4.5 / \phantom{0}0.4 & \textbf{39.1} / \textbf{14.2} & \textbf{\phantom{0}9.8} / \phantom{0}3.0 & \textbf{\phantom{0}4.5} / \textbf{\phantom{0}0.0}\\ \hline
        SN only  & 52.9 / 47.0 & 11.1 / 10.0 & \phantom{0}1.0 / \phantom{0}0.4 & 44.7 / 22.0 & 10.2 / \phantom{0}4.5 & \phantom{0}0.6 / \textbf{\phantom{0}0.1}\\
        SN + ST  & 51.9 / 47.3 & 11.5 / 10.1 & \phantom{0}1.4 / \phantom{0}0.6 & 45.6 / \textbf{26.3} & 10.6 / \textbf{\phantom{0}9.1} & \phantom{0}0.9 / \textbf{\phantom{0}0.1}\\
        \rowstyle{\color{blue}}
        SN + Dreaming  & \textbf{53.4} / \textbf{51.4} & \textbf{13.7} / \textbf{10.5} & \textbf{10.1} / \textbf{\phantom{0}9.1} & \textbf{50.0} / 24.3 & \textbf{11.3} / \textbf{\phantom{0}9.1} & \textbf{\phantom{0}9.1} / \textbf{\phantom{0}0.1}\\ \hline
    \end{tabular}
\end{subtable}

\begin{subtable}{\textwidth}
    \caption{\kitti to \waymo}
    \centering
    \begin{tabular}{=l|+c|+c|+c|+c|+c|+c}
        \multicolumn{1}{c|}{} & \multicolumn{3}{c|}{IoU 0.5} & \multicolumn{3}{c}{IoU 0.7} \\ \cline{2-7}
        \multicolumn{1}{c|}{\multirow{-2}{*}{\diagbox{Method}{Range(m)}}}            & \rowstyle{\color{black}} 0-30    & 30-50    & 50-80   & 0-30    & 30-50    & 50-80   \\ \hline
        \rowstyle{\color{gray}} in-domain  & 81.6 / 81.5 & 71.7 / 71.2 & 58.2 / 50.8 & 80.8 / 68.8 & 69.1 / 54.9 & 47.0 / 27.2\\ \hline
        no retraining  & 80.5 / 71.5 & 69.0 / 60.3 & 42.8 / 33.4 & 43.5 / 16.6 & 34.6 / 11.0 & 20.6 / \textbf{10.8} \\
        ST  & 81.0 / 78.4 & 69.5 / 61.4 & 48.5 / 39.3 & 48.2 / \textbf{18.6} & 36.8 / 16.0 & 23.6 / \phantom{0}7.0 \\
        \rowstyle{\color{blue}}
        Dreaming   & \textbf{81.1} / \textbf{78.5} & \textbf{69.9} / \textbf{61.8} & \textbf{50.2} / \textbf{41.0} & \textbf{51.4} / 13.8 & \textbf{44.5} / \textbf{16.7} & \textbf{25.6} / \phantom{0}7.8\\ \hline
        SN only  & 81.0 / 80.4 & 70.0 / 62.4 & 42.7 / 40.7 & 71.4 / \textbf{53.4} & 60.9 / 39.8 & 38.0 / 19.4\\
        SN + ST  & 81.1 / 80.6 & 70.5 / \textbf{68.5} & 49.7 / 42.5 & \textbf{78.1} / 53.1 & 61.7 / 45.1 & 40.2 / 20.6\\
        \rowstyle{\color{blue}}
        SN + Dreaming  & \textbf{81.2} / \textbf{80.8} & \textbf{70.9} / 68.3 & \textbf{51.1} / \textbf{48.3} & \textbf{78.1} / 51.6 & \textbf{67.4} / \textbf{45.5} & \textbf{41.1} / \textbf{20.8}\\ \hline
    \end{tabular}
\end{subtable}

\end{table*}

\begin{table*}[t]
\tabcolsep 4pt
\renewcommand{\arraystretch}{1.05}
\centering
\ContinuedFloat

\begin{subtable}{\textwidth}
    \caption{\argo to \kitti}
    \centering
    \begin{tabular}{=l|+c|+c|+c|+c|+c|+c}
        \multicolumn{1}{c|}{} & \multicolumn{3}{c|}{IoU 0.5} & \multicolumn{3}{c}{IoU 0.7} \\ \cline{2-7}
        \multicolumn{1}{c|}{\multirow{-2}{*}{\diagbox{Method}{Range(m)}}}            & \rowstyle{\color{black}} 0-30    & 30-50    & 50-80   & 0-30    & 30-50    & 50-80   \\ \hline
        \rowstyle{\color{gray}} in-domain  & 90.0 / 89.9 & 81.0 / 79.9 & 40.4 / 36.3 & 89.0 / 78.0 & 70.3 / 51.5 & 26.6 / \phantom{0}9.8\\ \hline
        no retraining  & 89.4 / 88.8 & 71.0 / 65.2 & 18.0 / 13.9 & 72.6 / 47.8 & 35.8 / 14.6 & \phantom{0}4.9 / \phantom{0}3.0 \\
        ST  & \textbf{89.5} / \textbf{89.2} & 73.2 / 68.5 & 28.2 / 22.5 & 76.8 / 52.9 & 44.2 / 20.6 & 13.1 / \phantom{0}2.1 \\
        \rowstyle{\color{blue}}
        Dreaming   & 89.3 / \textbf{89.2} & \textbf{74.6} / \textbf{72.4} & \textbf{30.1} / \textbf{25.6} & \textbf{77.6} / \textbf{54.7} & \textbf{49.9} / \textbf{24.3} & \textbf{14.5} / \textbf{\phantom{0}3.4}\\ \hline
        SN only  & 89.3 / 88.2 & 69.6 / 65.4 & 14.6 / 13.3 & 83.8 / 59.1 & 53.5 / 27.2 & \phantom{0}9.3 / \phantom{0}3.6\\
        SN + ST  & \textbf{89.8} / 89.3 & 74.4 / 72.8 & 22.3 / 21.3 & \textbf{87.0} / 70.4 & 62.2 / 37.7 & 16.4 / \phantom{0}7.1\\
        \rowstyle{\color{blue}}
        SN + Dreaming  & \textbf{89.8} / \textbf{89.4} & \textbf{75.4} / \textbf{73.8} & \textbf{29.4} / \textbf{25.4} & \textbf{87.0} / \textbf{73.5} & \textbf{62.8} / \textbf{41.9} & \textbf{17.2} / \textbf{10.3}\\ \hline
    \end{tabular}
\end{subtable}

\begin{subtable}{\textwidth}
    \caption{\argo to \lyft}
    \centering
    \begin{tabular}{=l|+c|+c|+c|+c|+c|+c}
        \multicolumn{1}{c|}{} & \multicolumn{3}{c|}{IoU 0.5} & \multicolumn{3}{c}{IoU 0.7} \\ \cline{2-7}
        \multicolumn{1}{c|}{\multirow{-2}{*}{\diagbox{Method}{Range(m)}}}            & \rowstyle{\color{black}} 0-30    & 30-50    & 50-80   & 0-30    & 30-50    & 50-80   \\ \hline
        \rowstyle{\color{gray}} in-domain  & 89.2 / 88.9 & 78.4 / 77.9 & 67.2 / 65.5 & 88.4 / 79.0 & 74.8 / 54.6 & 54.7 / 27.3\\ \hline
        no retraining  & 79.9 / 79.8 & 68.5 / 67.4 & 46.3 / 42.9 & 77.2 / 54.8 & 57.6 / 32.0 & 31.8 / 12.4 \\
        ST  & 80.2 / \textbf{80.1} & 72.7 / 67.9 & 48.8 / 46.0 & 78.3 / 55.4 & 63.3 / 35.8 & 35.0 / 11.3 \\
        \rowstyle{\color{blue}}
        Dreaming   & \textbf{80.3} / \textbf{80.1} & \textbf{73.8} / \textbf{72.4} & \textbf{54.5} / \textbf{47.6} & \textbf{78.9} / \textbf{57.3} & \textbf{64.2} / \textbf{37.1} & \textbf{35.6} / \textbf{12.8}\\ \hline
        SN only  & 79.6 / 79.2 & 67.1 / 66.0 & 44.8 / 41.8 & 75.7 / 54.7 & 55.3 / 30.4 & 32.4 / \textbf{14.2}\\
        SN + ST  & 80.1 / 79.9 & 72.2 / 67.4 & 47.4 / 45.3 & 78.3 / 56.0 & 63.6 / 38.0 & 36.4 / 13.5\\
        \rowstyle{\color{blue}}
        SN + Dreaming  & \textbf{80.2} / \textbf{80.0} & \textbf{73.5} / \textbf{68.2} & \textbf{54.0} / \textbf{47.5} & \textbf{78.5} / \textbf{56.8} & \textbf{64.1} / \textbf{39.3} & \textbf{41.9} / 14.1\\ \hline
    \end{tabular}
\end{subtable}

\begin{subtable}{\textwidth}
    \caption{\argo to \nusc}
    \centering
    \begin{tabular}{=l|+c|+c|+c|+c|+c|+c}
        \multicolumn{1}{c|}{} & \multicolumn{3}{c|}{IoU 0.5} & \multicolumn{3}{c}{IoU 0.7} \\ \cline{2-7}
        \multicolumn{1}{c|}{\multirow{-2}{*}{\diagbox{Method}{Range(m)}}}            & \rowstyle{\color{black}} 0-30    & 30-50    & 50-80   & 0-30    & 30-50    & 50-80   \\ \hline
        \rowstyle{\color{gray}} in-domain  & 65.6 / 64.5 & 18.4 / 17.2 & \phantom{0}3.5 / \phantom{0}2.8 & 57.8 / 37.0 & 15.8 / \phantom{0}5.3 & \phantom{0}2.6 / \phantom{0}0.8\\ \hline
        no retraining  & \textbf{51.6} / \textbf{45.3} & \textbf{10.3} / \textbf{\phantom{0}9.1} & \phantom{0}0.6 / \phantom{0}0.1 & \textbf{43.3} / 17.9 & \textbf{\phantom{0}9.1} / \phantom{0}0.3 & \phantom{0}0.2 / \textbf{\phantom{0}0.1} \\
        ST  & 47.7 / 43.4 & \phantom{0}6.3 / \phantom{0}5.4 & \textbf{\phantom{0}9.1} / \phantom{0}0.2 & 41.5 / \textbf{18.4} & \phantom{0}5.6 / \textbf{\phantom{0}4.5} & \phantom{0}0.3 / \textbf{\phantom{0}0.1} \\
        \rowstyle{\color{blue}}
        Dreaming   & 48.1 / 43.5 & \phantom{0}8.0 / \phantom{0}4.5 & \phantom{0}3.0 / \textbf{\phantom{0}3.0} & 41.4 / \textbf{18.4} & \phantom{0}5.3 / \phantom{0}2.3 & \textbf{\phantom{0}3.0} / \phantom{0}0.0\\ \hline
        SN only  & 47.6 / 45.6 & \phantom{0}5.7 / \phantom{0}4.5 & \phantom{0}0.5 / \phantom{0}0.2 & 43.1 / 18.2 & \phantom{0}4.5 / \phantom{0}0.7 & \phantom{0}0.1 / \phantom{0}0.0\\
        SN + ST  & 49.1 / 44.5 & 11.0 / \phantom{0}4.2 & \textbf{\phantom{0}9.1} / \textbf{\phantom{0}9.1} & 42.7 / 22.4 & 10.4 / \phantom{0}1.8 & \textbf{\phantom{0}9.1} / \textbf{\phantom{0}9.1}\\
        \rowstyle{\color{blue}}
        SN + Dreaming  & \textbf{50.7} / \textbf{45.9} & \textbf{12.8} / \textbf{10.1} & \textbf{\phantom{0}9.1} / \textbf{\phantom{0}9.1} & \textbf{44.1} / \textbf{24.7} & \textbf{10.7} / \textbf{\phantom{0}9.1} & \textbf{\phantom{0}9.1} / \phantom{0}4.5\\ \hline
    \end{tabular}
\end{subtable}

\begin{subtable}{\textwidth}
    \caption{\argo to \waymo}
    \centering
    \begin{tabular}{=l|+c|+c|+c|+c|+c|+c}
        \multicolumn{1}{c|}{} & \multicolumn{3}{c|}{IoU 0.5} & \multicolumn{3}{c}{IoU 0.7} \\ \cline{2-7}
        \multicolumn{1}{c|}{\multirow{-2}{*}{\diagbox{Method}{Range(m)}}}            & \rowstyle{\color{black}} 0-30    & 30-50    & 50-80   & 0-30    & 30-50    & 50-80   \\ \hline
        \rowstyle{\color{gray}} in-domain  & 81.6 / 81.5 & 71.7 / 71.2 & 58.2 / 50.8 & 80.8 / 68.8 & 69.1 / 54.9 & 47.0 / 27.2\\ \hline
        no retraining  & 81.2 / 80.4 & 69.3 / 61.7 & 50.4 / 47.8 & 70.8 / 43.7 & 59.6 / 35.0 & 39.1 / 18.8 \\
        ST  & \textbf{81.3} / \textbf{80.7} & 69.9 / \textbf{67.1} & 50.9 / 48.2 & \textbf{78.0} / \textbf{50.0} & \textbf{60.8} / 36.0 & 40.7 / \textbf{20.2} \\
        \rowstyle{\color{blue}}
        Dreaming   & \textbf{81.3} / 80.6 & \textbf{70.0} / 67.0 & \textbf{56.0} / \textbf{49.1} & 77.3 / 43.4 & 60.5 / \textbf{36.2} & \textbf{40.9} / \textbf{20.2}\\ \hline
        SN only  & 81.3 / 80.7 & 69.3 / 61.6 & 49.7 / 47.4 & 71.0 / 51.8 & 59.1 / 34.3 & 38.3 / 15.0\\
        SN + ST  & \textbf{81.4} / \textbf{81.0} & 70.1 / \textbf{67.7} & 50.7 / 48.0 & \textbf{78.4} / \textbf{55.5} & \textbf{61.1} / 41.1 & 40.4 / \textbf{20.0}\\
        \rowstyle{\color{blue}}
        SN + Dreaming  & \textbf{81.4} / \textbf{81.0} & \textbf{70.4} / 67.5 & \textbf{56.1} / \textbf{49.5} & 77.9 / 53.6 & \textbf{61.1} / \textbf{41.8} & \textbf{44.9} / \textbf{20.0}\\ \hline
    \end{tabular}
\end{subtable}

\end{table*}

\begin{table*}[t]
\tabcolsep 4pt
\renewcommand{\arraystretch}{1.05}
\centering
\ContinuedFloat

\begin{subtable}{\textwidth}
    \caption{\lyft to \kitti}
    \centering
    \begin{tabular}{=l|+c|+c|+c|+c|+c|+c}
        \multicolumn{1}{c|}{} & \multicolumn{3}{c|}{IoU 0.5} & \multicolumn{3}{c}{IoU 0.7} \\ \cline{2-7}
        \multicolumn{1}{c|}{\multirow{-2}{*}{\diagbox{Method}{Range(m)}}}            & \rowstyle{\color{black}} 0-30    & 30-50    & 50-80   & 0-30    & 30-50    & 50-80   \\ \hline
        \rowstyle{\color{gray}} in-domain  & 90.0 / 89.9 & 81.0 / 79.9 & 40.4 / 36.3 & 89.0 / 78.0 & 70.3 / 51.5 & 26.6 / \phantom{0}9.8\\ \hline
        no retraining  & 88.9 / 88.6 & 67.7 / 64.9 & 26.1 / 20.4 & 65.2 / 38.7 & 36.1 / 12.9 & \phantom{0}8.5 / \phantom{0}2.2 \\
        ST  & 89.9 / 89.6 & 73.5 / 68.8 & 30.8 / 24.0 & 71.7 / 40.8 & 42.2 / \textbf{20.5} & 10.8 / \phantom{0}2.9 \\
        \rowstyle{\color{blue}}
        Dreaming   & \textbf{90.0} / \textbf{89.7} & \textbf{74.7} / \textbf{72.8} & \textbf{33.9} / \textbf{26.4} & \textbf{74.2} / \textbf{42.8} & \textbf{46.2} / 18.7 & \textbf{12.1} / \textbf{\phantom{0}3.6}\\ \hline
        SN only  & 89.5 / 89.5 & 67.4 / 66.5 & 28.7 / 24.9 & 88.0 / 76.0 & 57.2 / 33.8 & 20.1 / \phantom{0}6.4\\
        SN + ST  & 90.0 / 90.0 & 73.2 / 72.3 & 31.9 / 28.4 & 88.4 / 77.2 & 63.1 / 42.3 & 22.1 / 10.0\\
        \rowstyle{\color{blue}}
        SN + Dreaming  & \textbf{90.1} / \textbf{90.1} & \textbf{76.2} / \textbf{74.8} & \textbf{36.1} / \textbf{32.1} & \textbf{88.6} / \textbf{78.0} & \textbf{64.6} / \textbf{42.5} & \textbf{23.6} / \textbf{10.7}\\ \hline
    \end{tabular}
\end{subtable}

\begin{subtable}{\textwidth}
    \caption{\lyft to \argo}
    \centering
    \begin{tabular}{=l|+c|+c|+c|+c|+c|+c}
        \multicolumn{1}{c|}{} & \multicolumn{3}{c|}{IoU 0.5} & \multicolumn{3}{c}{IoU 0.7} \\ \cline{2-7}
        \multicolumn{1}{c|}{\multirow{-2}{*}{\diagbox{Method}{Range(m)}}}            & \rowstyle{\color{black}} 0-30    & 30-50    & 50-80   & 0-30    & 30-50    & 50-80   \\ \hline
        \rowstyle{\color{gray}} in-domain  & 88.6 / 85.7 & 68.5 / 66.4 & 37.8 / 30.0 & 76.5 / 53.2 & 56.6 / 30.4 & 20.2 / 10.1\\ \hline
        no retraining  & 86.4 / 78.6 & 58.2 / 55.3 & 29.3 / 25.3 & 72.5 / 40.1 & 44.6 / 17.3 & 17.0 / \textbf{\phantom{0}9.1} \\
        ST  & 86.7 / \textbf{84.2} & 63.6 / 57.0 & 30.7 / 26.7 & 74.8 / \textbf{42.3} & \textbf{50.4} / \textbf{19.2} & 18.1 / \phantom{0}4.5 \\
        \rowstyle{\color{blue}}
        Dreaming   & \textbf{87.3} / 84.1 & \textbf{64.9} / \textbf{61.3} & \textbf{35.2} / \textbf{27.3} & \textbf{75.1} / 41.9 & 49.7 / 18.7 & \textbf{18.6} / \textbf{\phantom{0}9.1}\\ \hline
        SN only  & 86.5 / 78.4 & 58.1 / 55.0 & 29.2 / 22.2 & 72.3 / 46.5 & 44.3 / 13.0 & 17.7 / \phantom{0}9.1\\
        SN + ST  & 86.4 / 78.6 & 63.8 / 56.9 & 30.6 / 26.9 & 74.8 / \textbf{48.6} & \textbf{50.8} / \textbf{21.4} & 18.9 / \textbf{10.1}\\
        \rowstyle{\color{blue}}
        SN + Dreaming  & \textbf{87.4} / \textbf{84.1} & \textbf{64.8} / \textbf{61.5} & \textbf{34.9} / \textbf{27.9} & \textbf{74.9} / 45.6 & 50.4 / 20.6 & \textbf{19.0} / \phantom{0}9.9\\ \hline
    \end{tabular}
\end{subtable}

\begin{subtable}{\textwidth}
    \caption{\lyft to \nusc}
    \centering
    \begin{tabular}{=l|+c|+c|+c|+c|+c|+c}
        \multicolumn{1}{c|}{} & \multicolumn{3}{c|}{IoU 0.5} & \multicolumn{3}{c}{IoU 0.7} \\ \cline{2-7}
        \multicolumn{1}{c|}{\multirow{-2}{*}{\diagbox{Method}{Range(m)}}}            & \rowstyle{\color{black}} 0-30    & 30-50    & 50-80   & 0-30    & 30-50    & 50-80   \\ \hline
        \rowstyle{\color{gray}} in-domain  & 65.6 / 64.5 & 18.4 / 17.2 & \phantom{0}3.5 / \phantom{0}2.8 & 57.8 / 37.0 & 15.8 / \phantom{0}5.3 & \phantom{0}2.6 / \phantom{0}0.8\\ \hline
        no retraining  & \textbf{53.2} / 47.1 & \phantom{0}9.4 / \phantom{0}3.8 & \phantom{0}4.5 / \phantom{0}0.4 & 45.9 / 24.0 & \phantom{0}7.3 / \phantom{0}0.9 & \phantom{0}4.5 / \textbf{\phantom{0}0.3} \\
        ST  & 51.6 / 46.2 & 13.0 / \textbf{10.4} & \textbf{\phantom{0}9.1} / \textbf{\phantom{0}9.1} & 45.2 / 25.8 & 11.6 / \textbf{\phantom{0}9.1} & \textbf{\phantom{0}9.1} / \phantom{0}0.1 \\
        \rowstyle{\color{blue}}
        Dreaming   & 52.8 / \textbf{50.4} & \textbf{14.9} / 10.2 & \phantom{0}6.0 / \phantom{0}1.1 & \textbf{49.6} / \textbf{26.7} & \textbf{11.8} / \textbf{\phantom{0}9.1} & \phantom{0}4.5 / \phantom{0}0.1\\ \hline
        SN only  & \textbf{53.5} / 47.0 & \phantom{0}9.0 / \phantom{0}3.6 & \phantom{0}3.0 / \phantom{0}0.4 & 45.7 / 24.1 & \phantom{0}6.4 / \phantom{0}2.3 & \phantom{0}1.0 / \phantom{0}0.0\\
        SN + ST  & 52.0 / 46.6 & 12.9 / \textbf{10.5} & \phantom{0}4.5 / \textbf{\phantom{0}4.5} & 45.6 / \textbf{25.4} & 11.4 / \textbf{\phantom{0}9.1} & \textbf{\phantom{0}4.5} / \textbf{\phantom{0}0.1}\\
        \rowstyle{\color{blue}}
        SN + Dreaming  & 52.3 / \textbf{49.8} & \textbf{15.1} / 10.2 & \textbf{\phantom{0}5.0} / \phantom{0}1.5 & \textbf{48.8} / 23.3 & \textbf{12.1} / \textbf{\phantom{0}9.1} & \phantom{0}3.0 / \phantom{0}0.0\\ \hline
    \end{tabular}
\end{subtable}

\begin{subtable}{\textwidth}
    \caption{\lyft to \waymo}
    \centering
    \begin{tabular}{=l|+c|+c|+c|+c|+c|+c}
        \multicolumn{1}{c|}{} & \multicolumn{3}{c|}{IoU 0.5} & \multicolumn{3}{c}{IoU 0.7} \\ \cline{2-7}
        \multicolumn{1}{c|}{\multirow{-2}{*}{\diagbox{Method}{Range(m)}}}            & \rowstyle{\color{black}} 0-30    & 30-50    & 50-80   & 0-30    & 30-50    & 50-80   \\ \hline
        \rowstyle{\color{gray}} in-domain  & 81.6 / 81.5 & 71.7 / 71.2 & 58.2 / 50.8 & 80.8 / 68.8 & 69.1 / 54.9 & 47.0 / 27.2\\ \hline
        no retraining  & \textbf{81.5} / 81.2 & 70.9 / 69.8 & 51.6 / 49.6 & \textbf{79.5} / 55.8 & 61.6 / 44.5 & 40.5 / 18.6 \\
        ST  & \textbf{81.5} / \textbf{81.3} & 71.1 / 69.8 & 51.9 / 50.6 & 79.3 / \textbf{56.1} & \textbf{67.5} / \textbf{46.0} & \textbf{46.3} / \textbf{23.8} \\
        \rowstyle{\color{blue}}
        Dreaming   & \textbf{81.5} / 81.2 & \textbf{71.2} / \textbf{70.2} & \textbf{56.9} / \textbf{50.8} & 79.1 / 55.7 & 67.2 / 45.5 & 46.0 / 21.2\\ \hline
        SN only  & \textbf{81.4} / \textbf{81.3} & 71.0 / 70.1 & 51.1 / 49.5 & 79.8 / \textbf{65.5} & 61.1 / 46.1 & 39.0 / 21.3\\
        SN + ST  & \textbf{81.4} / \textbf{81.3} & 71.2 / 70.2 & 51.9 / 50.5 & \textbf{80.3} / 65.0 & \textbf{67.5} / \textbf{47.5} & \textbf{45.5} / \textbf{25.1}\\
        \rowstyle{\color{blue}}
        SN + Dreaming  & \textbf{81.4} / \textbf{81.3} & \textbf{71.3} / \textbf{70.4} & \textbf{56.9} / \textbf{50.9} & 80.1 / 64.9 & 67.1 / \textbf{47.5} & 44.6 / 22.8\\ \hline
    \end{tabular}
\end{subtable}

\end{table*}

\begin{table*}[t]
\tabcolsep 4pt
\renewcommand{\arraystretch}{1.05}
\centering
\ContinuedFloat

\begin{subtable}{\textwidth}
    \caption{\nusc to \kitti}
    \centering
    \begin{tabular}{=l|+c|+c|+c|+c|+c|+c}
        \multicolumn{1}{c|}{} & \multicolumn{3}{c|}{IoU 0.5} & \multicolumn{3}{c}{IoU 0.7} \\ \cline{2-7}
        \multicolumn{1}{c|}{\multirow{-2}{*}{\diagbox{Method}{Range(m)}}}            & \rowstyle{\color{black}} 0-30    & 30-50    & 50-80   & 0-30    & 30-50    & 50-80   \\ \hline
        \rowstyle{\color{gray}} in-domain  & 90.0 / 89.9 & 81.0 / 79.9 & 40.4 / 36.3 & 89.0 / 78.0 & 70.3 / 51.5 & 26.6 / \phantom{0}9.8\\ \hline
        no retraining  & 68.5 / 57.2 & 46.1 / 33.5 & \phantom{0}9.6 / \phantom{0}5.5 & 35.7 / \phantom{0}5.3 & 12.7 / \phantom{0}1.4 & \phantom{0}2.8 / \textbf{\phantom{0}1.8} \\
        ST  & \textbf{87.8} / \textbf{78.9} & 64.8 / 51.0 & 16.4 / \phantom{0}8.7 & 45.9 / 10.9 & 21.4 / \phantom{0}2.5 & \phantom{0}3.4 / \phantom{0}0.6 \\
        \rowstyle{\color{blue}}
        Dreaming   & 87.6 / 78.8 & \textbf{69.3} / \textbf{56.3} & \textbf{21.7} / \textbf{13.4} & \textbf{46.4} / \textbf{14.5} & \textbf{25.7} / \textbf{10.2} & \textbf{\phantom{0}5.4} / \phantom{0}0.5\\ \hline
        SN only  & 78.8 / 78.3 & 48.9 / 46.6 & \phantom{0}6.0 / \phantom{0}5.5 & 74.9 / 40.9 & 37.7 / 11.4 & \phantom{0}4.7 / \phantom{0}1.1\\
        SN + ST  & \textbf{88.7} / \textbf{88.5} & 65.6 / 63.5 & 16.6 / 13.0 & \textbf{84.3} / \textbf{61.2} & 53.3 / 22.3 & \textbf{10.6} / \textbf{\phantom{0}1.8}\\
        \rowstyle{\color{blue}}
        SN + Dreaming  & 88.5 / 88.2 & \textbf{69.9} / \textbf{65.8} & \textbf{23.9} / \textbf{20.9} & 84.0 / 60.8 & \textbf{53.9} / \textbf{24.5} & \phantom{0}8.9 / \phantom{0}1.5\\ \hline
    \end{tabular}
\end{subtable}

\begin{subtable}{\textwidth}
    \caption{\nusc to \argo}
    \centering
    \begin{tabular}{=l|+c|+c|+c|+c|+c|+c}
        \multicolumn{1}{c|}{} & \multicolumn{3}{c|}{IoU 0.5} & \multicolumn{3}{c}{IoU 0.7} \\ \cline{2-7}
        \multicolumn{1}{c|}{\multirow{-2}{*}{\diagbox{Method}{Range(m)}}}            & \rowstyle{\color{black}} 0-30    & 30-50    & 50-80   & 0-30    & 30-50    & 50-80   \\ \hline
        \rowstyle{\color{gray}} in-domain  & 88.6 / 85.7 & 68.5 / 66.4 & 37.8 / 30.0 & 76.5 / 53.2 & 56.6 / 30.4 & 20.2 / 10.1\\ \hline
        no retraining  & 39.5 / 37.0 & 21.6 / 17.4 & \phantom{0}3.0 / \phantom{0}3.0 & 28.6 / \phantom{0}5.1 & 17.6 / \textbf{\phantom{0}9.1} & \phantom{0}3.0 / \phantom{0}0.1 \\
        ST  & 65.2 / 59.5 & 32.6 / 25.9 & 12.4 / \phantom{0}9.1 & \textbf{51.9} / \textbf{15.7} & 24.3 / \textbf{\phantom{0}9.1} & \phantom{0}9.1 / \textbf{\phantom{0}0.6} \\
        \rowstyle{\color{blue}}
        Dreaming   & \textbf{65.9} / \textbf{63.2} & \textbf{50.0} / \textbf{40.3} & \textbf{17.7} / \textbf{13.0} & 50.7 / 10.6 & \textbf{32.3} / \phantom{0}2.9 & \textbf{\phantom{0}9.6} / \textbf{\phantom{0}0.6}\\ \hline
        SN only  & 49.6 / 47.8 & 20.8 / 16.4 & \phantom{0}4.5 / \phantom{0}4.5 & 38.2 / \phantom{0}8.2 & 16.6 / \phantom{0}2.3 & \phantom{0}4.5 / \textbf{\phantom{0}1.8}\\
        SN + ST  & 56.7 / 55.5 & 23.3 / 18.0 & \phantom{0}4.7 / \phantom{0}3.4 & 50.3 / \textbf{13.3} & 17.4 / \textbf{\phantom{0}9.1} & \phantom{0}3.3 / \phantom{0}0.5\\
        \rowstyle{\color{blue}}
        SN + Dreaming  & \textbf{64.3} / \textbf{61.8} & \textbf{45.7} / \textbf{40.9} & \textbf{19.4} / \textbf{14.1} & \textbf{50.8} / 11.0 & \textbf{33.4} / \phantom{0}3.0 & \textbf{12.8} / \phantom{0}0.8\\ \hline
    \end{tabular}
\end{subtable}

\begin{subtable}{\textwidth}
    \caption{\nusc to \lyft}
    \centering
    \begin{tabular}{=l|+c|+c|+c|+c|+c|+c}
        \multicolumn{1}{c|}{} & \multicolumn{3}{c|}{IoU 0.5} & \multicolumn{3}{c}{IoU 0.7} \\ \cline{2-7}
        \multicolumn{1}{c|}{\multirow{-2}{*}{\diagbox{Method}{Range(m)}}}            & \rowstyle{\color{black}} 0-30    & 30-50    & 50-80   & 0-30    & 30-50    & 50-80   \\ \hline
        \rowstyle{\color{gray}} in-domain  & 89.2 / 88.9 & 78.4 / 77.9 & 67.2 / 65.5 & 88.4 / 79.0 & 74.8 / 54.6 & 54.7 / 27.3\\ \hline
        no retraining  & 52.3 / 51.0 & 41.1 / 37.9 & 22.9 / 15.1 & 49.5 / 17.9 & 36.8 / 10.5 & 15.1 / \phantom{0}9.1 \\
        ST  & \textbf{79.7} / 78.0 & 65.9 / 58.0 & 39.1 / 30.2 & \textbf{77.0} / 20.2 & 57.0 / \textbf{13.0} & 30.0 / \textbf{10.4} \\
        \rowstyle{\color{blue}}
        Dreaming   & 79.6 / \textbf{78.1} & \textbf{66.8} / \textbf{63.2} & \textbf{46.4} / \textbf{37.8} & \textbf{77.0} / \textbf{20.7} & \textbf{62.1} / 10.4 & \textbf{36.4} / \phantom{0}5.2\\ \hline
        SN only  & 62.3 / 61.7 & 41.6 / 39.9 & 16.7 / 15.4 & 60.1 / 26.8 & 32.1 / \phantom{0}5.7 & 15.6 / \phantom{0}9.1\\
        SN + ST  & 78.6 / 77.9 & 56.1 / 53.7 & 23.1 / 13.9 & \textbf{77.3} / \textbf{32.4} & 53.0 / 13.8 & 20.4 / \phantom{0}3.0\\
        \rowstyle{\color{blue}}
        SN + Dreaming  & \textbf{78.7} / \textbf{78.0} & \textbf{64.7} / \textbf{57.5} & \textbf{39.7} / \textbf{36.1} & \textbf{77.3} / 31.6 & \textbf{56.0} / \textbf{16.0} & \textbf{29.6} / \textbf{11.0}\\ \hline
    \end{tabular}
\end{subtable}

\begin{subtable}{\textwidth}
    \caption{\nusc to \waymo}
    \centering
    \begin{tabular}{=l|+c|+c|+c|+c|+c|+c}
        \multicolumn{1}{c|}{} & \multicolumn{3}{c|}{IoU 0.5} & \multicolumn{3}{c}{IoU 0.7} \\ \cline{2-7}
        \multicolumn{1}{c|}{\multirow{-2}{*}{\diagbox{Method}{Range(m)}}}            & \rowstyle{\color{black}} 0-30    & 30-50    & 50-80   & 0-30    & 30-50    & 50-80   \\ \hline
        \rowstyle{\color{gray}} in-domain  & 81.6 / 81.5 & 71.7 / 71.2 & 58.2 / 50.8 & 80.8 / 68.8 & 69.1 / 54.9 & 47.0 / 27.2\\ \hline
        no retraining  & 62.2 / 61.2 & 50.7 / 42.1 & 24.9 / 22.5 & 60.5 / 25.3 & 41.7 / 13.8 & 21.3 / \phantom{0}4.5 \\
        ST  & \textbf{80.7} / \textbf{71.9} & 68.7 / 60.9 & 42.5 / 39.6 & 71.6 / \textbf{34.5} & 60.7 / \textbf{24.6} & 37.0 / \phantom{0}8.8 \\
        \rowstyle{\color{blue}}
        Dreaming   & \textbf{80.7} / \textbf{71.9} & \textbf{69.4} / \textbf{61.3} & \textbf{50.1} / \textbf{45.5} & \textbf{71.7} / 33.5 & \textbf{61.2} / 24.3 & \textbf{39.3} / \textbf{10.7}\\ \hline
        SN only  & 71.1 / 70.3 & 50.6 / 41.8 & 25.1 / 16.6 & 61.1 / 34.8 & 40.5 / 14.4 & 16.1 / \phantom{0}2.8\\
        SN + ST  & 80.7 / 79.1 & 67.4 / 59.2 & 41.3 / 32.1 & 71.4 / \textbf{44.0} & 58.2 / 27.3 & 29.7 / \phantom{0}7.4\\
        \rowstyle{\color{blue}}
        SN + Dreaming  & \textbf{81.1} / \textbf{79.2} & \textbf{69.5} / \textbf{61.5} & \textbf{50.1} / \textbf{41.3} & \textbf{71.7} / 42.2 & \textbf{60.7} / \textbf{27.6} & \textbf{38.5} / \textbf{14.4}\\ \hline
    \end{tabular}
\end{subtable}

\end{table*}

\begin{table*}[t]
\tabcolsep 4pt
\renewcommand{\arraystretch}{1.05}
\centering
\ContinuedFloat

\begin{subtable}{\textwidth}
    \caption{\waymo to \kitti}
    \centering
    \begin{tabular}{=l|+c|+c|+c|+c|+c|+c}
        \multicolumn{1}{c|}{} & \multicolumn{3}{c|}{IoU 0.5} & \multicolumn{3}{c}{IoU 0.7} \\ \cline{2-7}
        \multicolumn{1}{c|}{\multirow{-2}{*}{\diagbox{Method}{Range(m)}}}            & \rowstyle{\color{black}} 0-30    & 30-50    & 50-80   & 0-30    & 30-50    & 50-80   \\ \hline
        \rowstyle{\color{gray}} in-domain  & 90.0 / 89.9 & 81.0 / 79.9 & 40.4 / 36.3 & 89.0 / 78.0 & 70.3 / 51.5 & 26.6 / \phantom{0}9.8\\ \hline
        no retraining  & 88.4 / 87.0 & 64.7 / 55.2 & 14.3 / \phantom{0}8.5 & 45.8 / 10.4 & 22.9 / \phantom{0}3.1 & \phantom{0}2.5 / \phantom{0}0.8 \\
        ST  & 89.3 / 88.3 & 70.5 / 62.8 & 25.6 / 18.1 & \textbf{50.7} / 10.1 & 28.6 / 11.9 & \phantom{0}4.3 / \phantom{0}2.3 \\
        \rowstyle{\color{blue}}
        Dreaming   & \textbf{89.5} / \textbf{88.7} & \textbf{72.6} / \textbf{65.8} & \textbf{27.8} / \textbf{19.7} & 49.8 / \textbf{16.3} & \textbf{31.9} / \textbf{12.3} & \textbf{\phantom{0}5.0} / \textbf{\phantom{0}3.0}\\ \hline
        SN only  & 88.6 / 88.5 & 63.0 / 61.9 & 11.6 / 10.7 & 84.1 / 53.8 & 51.7 / 28.3 & \phantom{0}8.4 / \phantom{0}5.1\\
        SN + ST  & \textbf{89.7} / \textbf{89.7} & 72.5 / 69.0 & 19.8 / 18.6 & \textbf{87.1} / 60.6 & \textbf{60.8} / \textbf{38.6} & 13.8 / \textbf{\phantom{0}6.2}\\
        \rowstyle{\color{blue}}
        SN + Dreaming  & 89.6 / 89.6 & \textbf{73.5} / \textbf{72.3} & \textbf{22.1} / \textbf{19.6} & 86.3 / \textbf{63.1} & 58.0 / 36.4 & \textbf{14.0} / \phantom{0}4.0\\ \hline
    \end{tabular}
\end{subtable}

\begin{subtable}{\textwidth}
    \caption{\waymo to \argo}
    \centering
    \begin{tabular}{=l|+c|+c|+c|+c|+c|+c}
        \multicolumn{1}{c|}{} & \multicolumn{3}{c|}{IoU 0.5} & \multicolumn{3}{c}{IoU 0.7} \\ \cline{2-7}
        \multicolumn{1}{c|}{\multirow{-2}{*}{\diagbox{Method}{Range(m)}}}            & \rowstyle{\color{black}} 0-30    & 30-50    & 50-80   & 0-30    & 30-50    & 50-80   \\ \hline
        \rowstyle{\color{gray}} in-domain  & 88.6 / 85.7 & 68.5 / 66.4 & 37.8 / 30.0 & 76.5 / 53.2 & 56.6 / 30.4 & 20.2 / 10.1\\ \hline
        no retraining  & 84.0 / 76.3 & 54.4 / 51.3 & 24.7 / 19.6 & 69.6 / 28.4 & 40.6 / 13.8 & 15.9 / \phantom{0}1.6 \\
        ST  & 85.7 / \textbf{78.6} & 55.5 / 52.9 & 23.4 / 21.3 & \textbf{70.7} / \textbf{29.8} & \textbf{44.3} / \textbf{14.0} & 14.9 / \phantom{0}2.1 \\
        \rowstyle{\color{blue}}
        Dreaming   & \textbf{85.8} / 78.3 & \textbf{56.7} / \textbf{54.6} & \textbf{28.3} / \textbf{22.3} & 64.8 / 28.8 & 44.1 / 13.5 & \textbf{17.3} / \textbf{\phantom{0}3.0}\\ \hline
        SN only  & 83.3 / 74.9 & 54.3 / 47.2 & 19.2 / 16.3 & 69.3 / 29.2 & 43.2 / 16.9 & 10.3 / \phantom{0}3.0\\
        SN + ST  & 85.2 / 78.0 & 55.4 / 52.8 & 22.8 / 20.4 & \textbf{73.9} / \textbf{37.6} & \textbf{45.8} / \textbf{19.8} & 14.8 / \textbf{\phantom{0}9.1}\\
        \rowstyle{\color{blue}}
        SN + Dreaming  & \textbf{85.4} / \textbf{78.1} & \textbf{56.0} / \textbf{53.6} & \textbf{27.7} / \textbf{22.1} & 71.8 / 35.2 & 45.5 / 16.8 & \textbf{17.6} / \textbf{\phantom{0}9.1}\\ \hline
    \end{tabular}
\end{subtable}

\begin{subtable}{\textwidth}
    \caption{\waymo to \lyft}
    \centering
    \begin{tabular}{=l|+c|+c|+c|+c|+c|+c}
        \multicolumn{1}{c|}{} & \multicolumn{3}{c|}{IoU 0.5} & \multicolumn{3}{c}{IoU 0.7} \\ \cline{2-7}
        \multicolumn{1}{c|}{\multirow{-2}{*}{\diagbox{Method}{Range(m)}}}            & \rowstyle{\color{black}} 0-30    & 30-50    & 50-80   & 0-30    & 30-50    & 50-80   \\ \hline
        \rowstyle{\color{gray}} in-domain  & 89.2 / 88.9 & 78.4 / 77.9 & 67.2 / 65.5 & 88.4 / 79.0 & 74.8 / 54.6 & 54.7 / 27.3\\ \hline
        no retraining  & 80.8 / \textbf{80.8} & 67.8 / 66.8 & 45.3 / 43.0 & 78.3 / 55.3 & 56.7 / 31.2 & 34.4 / 10.5 \\
        ST  & 80.7 / 80.6 & 74.6 / 73.4 & 54.0 / 48.1 & 78.9 / 54.9 & \textbf{65.4} / \textbf{32.8} & 42.9 / 17.2 \\
        \rowstyle{\color{blue}}
        Dreaming   & \textbf{86.8} / \textbf{80.8} & \textbf{75.4} / \textbf{74.0} & \textbf{55.5} / \textbf{49.6} & \textbf{79.2} / \textbf{55.5} & \textbf{65.4} / 32.6 & \textbf{44.6} / \textbf{17.3}\\ \hline
        SN only  & 80.5 / 80.2 & 66.4 / 65.1 & 38.4 / 37.1 & 78.0 / 55.9 & 55.8 / 31.4 & 33.5 / 15.4\\
        SN + ST  & 86.8 / 80.7 & 74.2 / 73.3 & 45.5 / 44.7 & 79.8 / 63.9 & 65.4 / \textbf{38.1} & 36.1 / 14.5\\
        \rowstyle{\color{blue}}
        SN + Dreaming  & \textbf{87.3} / \textbf{81.1} & \textbf{75.3} / \textbf{73.8} & \textbf{48.0} / \textbf{47.0} & \textbf{80.3} / \textbf{64.6} & \textbf{66.3} / \textbf{38.1} & \textbf{38.4} / \textbf{18.4}\\ \hline
    \end{tabular}
\end{subtable}

\begin{subtable}{\textwidth}
    \caption{\waymo to \nusc}
    \centering
    \begin{tabular}{=l|+c|+c|+c|+c|+c|+c}
        \multicolumn{1}{c|}{} & \multicolumn{3}{c|}{IoU 0.5} & \multicolumn{3}{c}{IoU 0.7} \\ \cline{2-7}
        \multicolumn{1}{c|}{\multirow{-2}{*}{\diagbox{Method}{Range(m)}}}            & \rowstyle{\color{black}} 0-30    & 30-50    & 50-80   & 0-30    & 30-50    & 50-80   \\ \hline
        \rowstyle{\color{gray}} in-domain  & 65.6 / 64.5 & 18.4 / 17.2 & \phantom{0}3.5 / \phantom{0}2.8 & 57.8 / 37.0 & 15.8 / \phantom{0}5.3 & \phantom{0}2.6 / \phantom{0}0.8\\ \hline
        no retraining  & 46.4 / 43.2 & \phantom{0}3.0 / \phantom{0}3.0 & \phantom{0}0.2 / \phantom{0}0.0 & 42.6 / 23.9 & \phantom{0}3.0 / \phantom{0}0.2 & \phantom{0}0.1 / \phantom{0}0.0 \\
        ST  & 51.4 / 46.5 & 10.7 / \phantom{0}9.1 & \phantom{0}3.0 / \phantom{0}0.3 & 45.8 / \textbf{26.4} & \phantom{0}9.1 / \textbf{\phantom{0}3.0} & \phantom{0}3.0 / \phantom{0}0.0 \\
        \rowstyle{\color{blue}}
        Dreaming   & \textbf{53.8} / \textbf{48.0} & \textbf{12.4} / \textbf{10.0} & \textbf{\phantom{0}9.1} / \textbf{\phantom{0}9.1} & \textbf{47.4} / 23.7 & \textbf{11.1} / \textbf{\phantom{0}3.0} & \textbf{\phantom{0}9.1} / \textbf{\phantom{0}0.8}\\ \hline
        SN only  & 45.8 / 43.5 & \phantom{0}9.1 / \phantom{0}9.1 & \phantom{0}0.1 / \phantom{0}0.0 & 42.9 / 23.5 & \phantom{0}9.1 / \phantom{0}0.2 & \phantom{0}0.0 / \textbf{\phantom{0}0.0}\\
        SN + ST  & 51.0 / 45.8 & \phantom{0}9.1 / \phantom{0}9.1 & \phantom{0}1.0 / \phantom{0}0.3 & 44.7 / 25.6 & \phantom{0}9.1 / \textbf{\phantom{0}9.1} & \phantom{0}1.0 / \textbf{\phantom{0}0.0}\\
        \rowstyle{\color{blue}}
        SN + Dreaming  & \textbf{53.7} / \textbf{47.6} & \textbf{12.4} / \textbf{10.2} & \textbf{\phantom{0}2.8} / \textbf{\phantom{0}0.8} & \textbf{46.8} / \textbf{26.5} & \textbf{11.2} / \phantom{0}2.3 & \textbf{\phantom{0}2.3} / \textbf{\phantom{0}0.0}\\ \hline
    \end{tabular}
\end{subtable}

\end{table*}
\clearpage

\end{document}